%% file: main.tex
\documentclass{article} 
\usepackage{main,times}

\input{math_commands.tex}

\input{preamble}
\usepackage{hyperref}
\usepackage{url}
\usepackage{graphicx} 
\usepackage{rotating}
\usepackage{subfig}   
\usepackage{booktabs}
\usepackage{threeparttable}
\usepackage{makecell}
\usepackage{changepage}

\title{
ARTDECO: Towards Efficient and High-Fidelity On-the-Fly 3D Reconstruction with Structured Scene Representation
}

\author{Guanghao Li$^{1,2,3}$\thanks{Equal contribution} \quad
Kerui Ren$^{1,4}$\protect\footnotemark[1] \quad
Linning Xu$^{1,5}$ \quad 
Zhewen Zheng$^{1,6}$ \quad  \\
\textbf{Changjian Jiang}$^{1,7}$ \quad 
\textbf{Xin Gao}$^{1,2}$ \quad
\textbf{Bo Dai}$^{8}$ \quad
\textbf{Jian Pu}$^{2}$\thanks{Corresponding author.} \quad
\textbf{Mulin Yu}$^{1}$\protect\footnotemark[2] \quad
\textbf{Jiangmiao Pang}$^{1}$ \quad \\
{\small$^1$Shanghai Artificial Intelligence Laboratory, $^2$Fudan University, $^3$Shanghai Innovation Institute,} \\
{\small$^4$Shanghai Jiao Tong University, $^5$The Chinese University of Hong Kong, $^6$Carnegie Mellon University, } \\
{\small $^7$Zhejiang University, $^8$The University of Hong Kong}\\
}

\iclrfinalcopy 
\begin{document}

\maketitle

\begin{figure*}[htbp]
\vspace{-1cm}
\includegraphics[width=\linewidth]{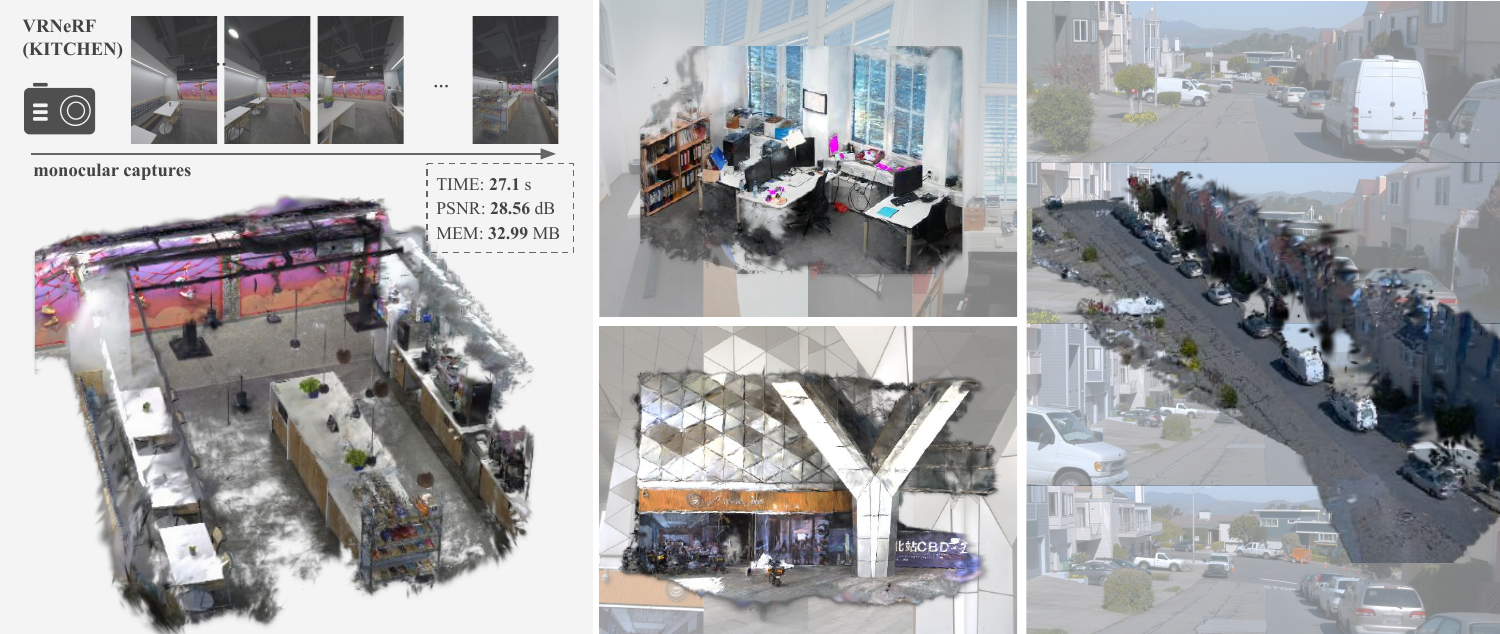}
\caption{ARTDECO delivers high-fidelity, interactive 3D reconstruction from monocular images, combining efficiency with robustness across indoor and outdoor scenes.
}
\label{fig:teaser}
\end{figure*}

\input{sec/abstract}
\input{sec/introduction}
\input{sec/related_work}
\input{sec/method}
\input{sec/experiments}
\input{sec/limitation}
\input{sec/conclusion}

\bibliography{main}
\bibliographystyle{main}

\appendix
\input{sec/supplementary}

\end{document}

%% file: math_commands.tex
\usepackage{amsmath,amsfonts,bm}

\def\eqref#1{equation~\ref{#1}}

\def\1{\bm{1}}

\DeclareMathAlphabet{\mathsfit}{\encodingdefault}{\sfdefault}{m}{sl}
\SetMathAlphabet{\mathsfit}{bold}{\encodingdefault}{\sfdefault}{bx}{n}



%% file: preamble.tex
\definecolor{Coral}{rgb}{1, 0.47, 0.24}

\usepackage{booktabs}
\usepackage{multirow}
\usepackage{graphicx}
\usepackage{array}
\usepackage{gensymb}  
\usepackage{fancyvrb}
\usepackage{lipsum}
\usepackage{xspace}
\usepackage{pifont}

%% file: sec/abstract.tex
\begin{abstract}
On-the-fly 3D reconstruction from monocular image sequences is a long-standing challenge in computer vision, critical for applications such as real-to-sim, AR/VR, and robotics. Existing methods face a major tradeoff: per-scene optimization yields high fidelity but is computationally expensive, whereas feed-forward foundation models enable real-time inference but struggle with accuracy and robustness.
In this work, we propose ARTDECO, a unified framework that combines the efficiency of feed-forward models with the reliability of SLAM-based pipelines. 
ARTDECO uses 3D foundation models for pose estimation and point prediction, coupled with a Gaussian decoder that transforms multi-scale features into structured 3D Gaussians.
To sustain both fidelity and efficiency at scale, we design a hierarchical Gaussian representation with a LoD-aware rendering strategy, which improves rendering fidelity while reducing redundancy.
Experiments on eight diverse indoor and outdoor benchmarks show that ARTDECO delivers interactive performance comparable to SLAM, robustness similar to feed-forward systems, and reconstruction quality close to per-scene optimization, providing a practical path toward on-the-fly digitization of real-world environments with both accurate geometry and high visual fidelity. Explore more demos on our project page: \href{https://city-super.github.io/artdeco/}{\textcolor{magenta}{\textbf{https://city-super.github.io/artdeco/}}}.     
\end{abstract}

%% file: sec/introduction.tex
\section{Introduction}
\label{sec:intro}
High-fidelity 3D reconstruction from monocular image sequences is a long-standing goal in computer vision. 
Monocular data are relatively inexpensive, ubiquitous, and easy to capture, while accurate 3D scene representations are crucial for downstream applications such as embodied intelligence, AR/VR, or real-to-sim content creation.
Recently, 3D Gaussian Splatting~\citep{3dgsOriginal} has emerged as an efficient scene representation with strong empirical results. 
However, in monocular settings the lack of reliable geometric cues such as ambiguous scale, limited parallax, motion blur, and poor overlap makes it difficult to achieve accuracy, speed, and robustness at the same time.
As a consequence, many existing systems optimize one objective at the expense of the others, limiting their practicality for online deployment.

Current 3DGS-based 3D reconstruction methods fall into two paradigms. Per-scene optimization methods~\citep{matsuki2023gaussian,huang2023photo,meulemanOntheflyNVS2025} rely on image poses estimated from Structure from Motion (SfM) or Simultaneous Localization and Mapping (SLAM), achieving high accuracy but at the cost of substantial computation, with robustness limited by the fragility of these pipelines. In contrast, recent feed-forward models~\citep{tang2024mv, ye2024noposplat, jiangAnySplat2025, zhang2025flare} leverage large-scale data to learn monocular priors and directly regress poses and Gaussian primitives with attention, enabling fast and robust inference across diverse scenes but with limited rendering fidelity and weak consistency. This tradeoff highlights the need for approaches that combine the efficiency of feed-forward models with the strengths of per-scene optimization methods to deliver accurate, robust, real-time reconstruction.
Beyond the efficiency–accuracy tradeoff, 3DGS pipelines are also highly sensitive to scene scale. As the scene grows, the number of Gaussian primitives required for training and rendering increases rapidly, which reduces efficiency. Prior attempts to address this either apply post-hoc anchor-based pruning, which lowers computation but introduces boundary artifacts and increases memory cost, or add multi-scale Gaussians during training, which mitigates artifacts but lacks explicit structural organization. These limitations underscore the need for a principled and practical level-of-detail (LoD) mechanism in 3DGS.

\textbf{ARTDECO}\footnote{Beyond the acronym, the name also evokes the \emph{Art Deco movement}, valued for structure, geometry, and clarity of form. This metaphor reflects our system’s emphasis on structured scene representations.} derives its name from a streamlined pipeline that unifies \textbf{A}ccurate localization, \textbf{R}obust recons\textbf{t}ruction, and \textbf{Deco}der-based rendering, with the aim of enabling on-the-fly 3D scene reconstruction and rendering.
The core of ARTDECO is the goal of balancing real-time performance, accuracy, and robustness. It employs feed-forward models as data priors to reduce monocular ambiguities while maintaining the efficiency required for interactive use. To address the global inconsistency often seen in feed-forward approaches, ARTDECO integrates loop detection with lightweight bundle adjustment. Finally, a hierarchical semi-implicit Gaussian structure with LoD-aware densification provides level-of-detail control, helping the system scale without excessive loss of fidelity or efficiency. Together, these components support practical real-time 3D reconstruction across diverse indoor and outdoor settings.

Our main contributions can be summarized as follows:
\begin{adjustwidth}{0.1em}{0pt}

\begin{itemize}
    \item We present ARTDECO, an integrated system that unifies localization, reconstruction, and rendering into a single pipeline, designed to operate robustly across various environments. 
    \item Notably, we incorporate \emph{feed-forward foundation models} as modular components for pose estimation, loop closure detection, and dense point prediction. This integration improves localization accuracy and mapping stability while preserving efficiency.
    \item We further propose a \emph{hierarchical semi-implicit Gaussian representation} with a LoD-aware densification strategy, enabling a principled trade-off between reconstruction fidelity and rendering efficiency, critical for large-scale, navigable environments.  
    \item Extensive indoor and outdoor experiments show that ARTDECO achieves SLAM-level efficiency, feed-forward robustness, and near per-scene optimization quality, validating its effectiveness for practical on-the-fly 3D digitization.
\end{itemize}
\end{adjustwidth}

%% file: sec/related_work.tex
\section{Related work}
\label{sec:related_work}

\subsection{Multi-view reconstruction and rendering}
Neural Radiance Fields (NeRF)~\citep{mildenhallNerf2021} have attracted significant attention in novel view synthesis (NVS). NeRF and its variants~\citep{barronMipnerf2021,barronMipnerf3602022,barronZipnerf2023,zhangNerfpp2020,verbinRefnerf2022} model continuous volumetric fields and achieve high-quality image synthesis. However, the reliance on expensive volume rendering and large networks results in long training times and hinders real-time applications. To address these limitations, several works~\citep{mullerInstantNGP2022, Xu_2022_CVPR, Sun_2022_CVPR} accelerate both training and rendering by introducing hybrid or explicit scene representations. Recently, 3D Gaussian Splatting (3DGS)~\citep{kerbl3DGS2023} has shown remarkable progress in high-fidelity reconstruction and real-time rendering by representing scenes with anisotropic Gaussians and leveraging an efficient tile-based rasterizer. Following its introduction, research has largely focused on model compression~\citep{fan2024lightgaussian, fcgs2025}, large-scale scene processing~\citep{ren2024octree, Jiang_2025_CVPR, hierarchicalgaussians24}, and geometry reconstruction~\citep{Huang2DGS2024, Yu2024GOF, guedon2025milo, yu2024gsdf}.
Despite these advances in novel view synthesis (NVS), most methods assume access to accurate camera poses, typically estimated via Structure-from-Motion (SfM)~\citep{schonbergerCOLMAP2016, schonbergerPixelwiseViewSelection2016, panGLOMAP2024}, which imposes considerable preprocessing costs for large-scale or in-the-wild captures. To alleviate this reliance, several works~\citep{linBarf2021, fuColmapfree3dGaussian2024, linLongSplat2025} propose joint optimization of camera poses and scene parameters. However, these approaches remain computationally intensive, are sensitive to wide-baseline settings, or still depend on costly post-refinement.

\subsection{Streaming per-scene reconstruction}
Classical visual SLAM systems~\citep{mur-artalOrbslam22017, engel2017direct, teed2023deep, li2025papl} provide online tracking, mapping, and loop closure, but they fall short in producing high-fidelity maps. To overcome this limitation, recent works integrate volumetric rendering techniques into SLAM to enable online NVS. Among these works, NeRF-based SLAM methods~\citep{sucar2021imap, zhuNICESLAM2022, zhang2023go, li2026ec} exhibit photorealistic reconstruction but remain computationally expensive due to per-ray volumetric rendering. 
By contrast, 3DGS has gained traction for SLAM integration thanks to its explicit representation and efficient rendering. Utilizing the differential pipeline of 3DGS, some studies~\citep{matsuki2023gaussian,cgSLAMHu, keetha2023splatam,yanGSSLAM2024,COMPACTSLAMdeng,yu2025rgb,li2025constrained,cheng2025outdoor,lin2025longsplat} directly propagate the gradient from the rendering loss to pose, while others~\citep{yugay2023gaussian, huang2023photo, RTGSLAMPeng, gsicpSLAMHa, QSLAMPeng, tianSEGSSLAM, sandstrom2025splat} leverage traditional SLAM Modules to provide accurate pose. 
However, in monocular setting these systems often struggle to balance robustness, accuracy, and runtime efficiency. Recently, On-the-fly NVS~\citep{meulemanOntheflyNVS2025} has shown that GPU-friendly mini-bundle adjustment with incremental 3DGS updates can enable interactive reconstruction, though robustness on casual unposed videos remains limited.

\subsection{Feed-forward Models}
Pretrained on large-scale datasets, recent feed-forward models~\citep{wang2025vggt,dust3r_cvpr24} reconstruct 3D scenes directly, avoiding per-scene optimization. These approaches can be divided into pose-aware and pose-free methods. Pose-aware models take images together with camera poses as input, enabling rapid reconstruction~\citep{charatan23pixelsplat, chen2024mvsplat, gslrm2024, jiang2025rayzer}. Pose-free models, in contrast, perform fully end-to-end reconstruction from raw images alone, typically representing scenes with either point maps~\citep{dust3r_cvpr24, mast3r_eccv24, wang2025vggt, wang2025pi3, muraiMASt3RSLAM2024} or 3DGS~\citep{jiangAnySplat2025}. 
Notably, these feed-forward methods offer robustness across diverse scenarios, remove the need for preprocessing, and allow fast inference suitable for interactive use. However, they generally achieve lower accuracy than per-scene optimized methods, and face challenges with maintaining global consistency, handling high-resolution inputs, and processing long sequences.

%% file: sec/method.tex
\section{Method}
\label{sec:method}

\begin{figure*}[t!]
\includegraphics[width=\linewidth]{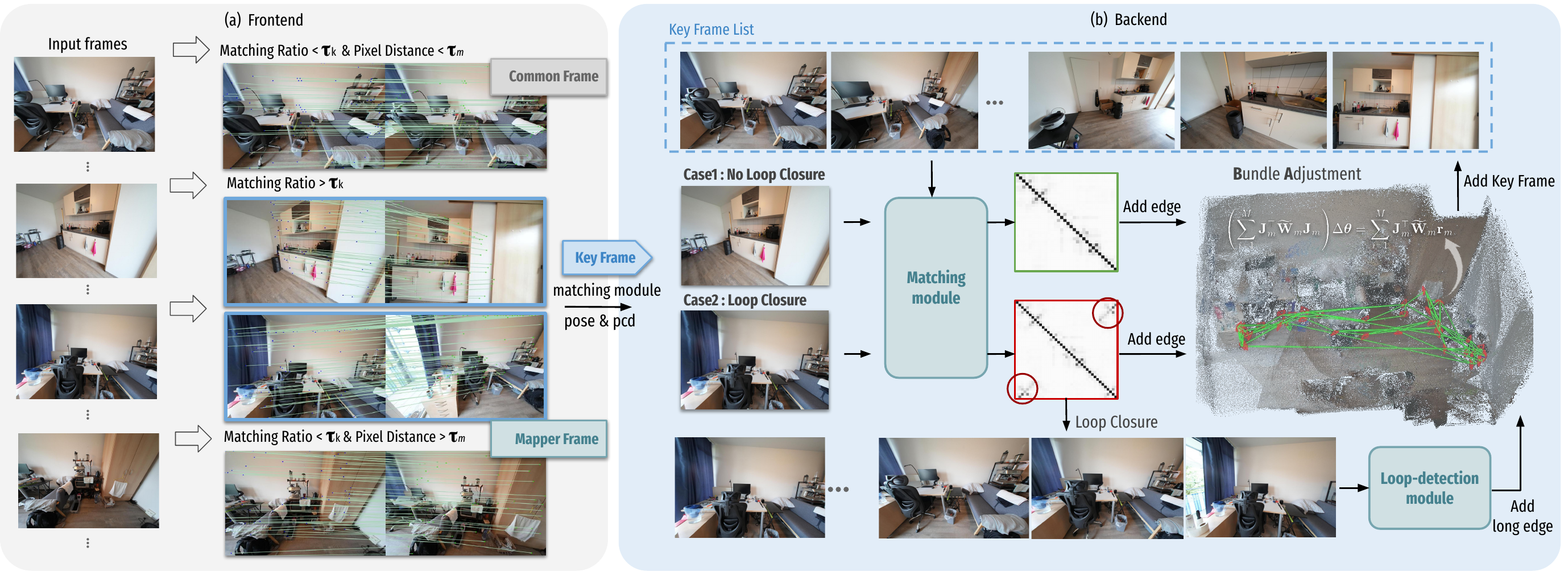}
\caption{
\textbf{Frontend and backend modules.}
(a) \emph{Frontend}: Images are captured from the scene and streamed into the front-end part. 
Each incoming frame is aligned with the latest keyframe using a \emph{matching module} to compute pixel correspondences.
Based on the correspondence ratio and pixel displacement, the frame is classified as a keyframe, a mapper frame, or a common frame. The selected frame, along with its pose and point cloud, is then passed to the back-end.
(b) \emph{Backend}: 
For each new keyframe, a \emph{loop-detection module} evaluates its similarity with previous keyframes. If a loop is detected, the most relevant candidates are refined and connected in the factor graph; otherwise, the keyframe is linked only to recent frames. Finally, global pose optimization is performed with Gauss–Newton, and other frames are adjusted accordingly.
We instantiate the matching module with MASt3R~\citep{mast3r_eccv24} and the loop-detection module with $\pi^3$~\citep{wang2025pi3}.
}
\vspace{-12pt}
\label{fig:slam}
\end{figure*}

We aim to recover a high-fidelity static 3D scene together with the corresponding camera poses from a monocular image sequence. 
Given a sequence of monocular RGB frames $\{ \mathbf{I}_i\}_{i=1}^{N}$, with or without known camera intrinsics $\mathbf{K} \in \mathbb{R}^{3 \times 3}$, we estimate the camera poses $\{ \mathbf{R}_i \mid \mathbf{t}_i \}_{i=1}^{N}$ associated with each image, as well as a set of Gaussian primitives $\{ \mathcal{G}_j \}_{j=1}^{M}$ that compactly represent the 3D scene. 
By default, we assume the scene is static and rigid, and that all geometric information is inferred purely from monocular cues without external sensors.

As illustrated in Fig.~\ref{fig:slam} and~\ref{fig:mapping}, ARTDECO processes the sequence in a streaming SLAM-style pipeline consisting of three modules: \textit{frontend}, \textit{backend}, and \textit{mapping}. (1) The frontend estimates relative poses and categorizes frames into common, mapping, or keyframes (Sec.~\ref{section:fronted}). 
(2) The backend refines keyframe poses through loop closure and global bundle adjustment (Sec.~\ref{section:Backend}). (3) Finally, image-wise pointmaps initialize 3D Gaussians, which are incrementally optimized in the mapping module (Sec.~\ref{sec:mapping}).

\subsection{Frontend Module}
\label{section:fronted}
For each input frame, the frontend estimates its pose relative to the latest keyframe and categorizes it as \emph{common}, \emph{mapping}, or \emph{keyframe}. We assume a pinhole camera with fixed intrinsics and a shared optical center. If the focal length is unknown, it is initialized from the first $k_f$ GeoCalib~\citep{veicht2024geocalib} estimates and jointly refined during pose estimation.

\noindent \textbf{Pose Estimation.} 
MASt3R~\citep{leroyMASt3R2024} serves as our \emph{matching module}, a two-view reconstruction and matching prior, to improve camera tracking and focal length estimation. Following MASt3R-SLAM~\citep{muraiMASt3RSLAM2024}, we obtain frame-wise pointmaps, their confidence scores, and pixel correspondences between the current frame and the latest keyframe. The 3D points from the current frame are projected into the keyframe image plane, and the relative pose $\mathbf{T}_{KC} \in \mathrm{SIM}(3)$ is estimated by minimizing reprojection residuals with a Gauss–Newton solver. 
Since MASt3R predictions are less stable near object boundaries, we weight residuals by per-point uncertainty. For each point $\mathbf{x}_c$ in the current frame, we estimate a local covariance $\boldsymbol{\Sigma}_c \in \mathbb{R}^{3\times 3}$ from neighbors within radius $\delta$. We then project $\boldsymbol{\Sigma}_c$ to the current keyframe's measurement space, which is used to filter out unreliable re-projection residuals. Besides, if the focal length is not provided, it is jointly optimized along with the relative pose. Further derivations are given in \ref{sec:jacobians}.

\noindent \textbf{Keyframe Selection.} 
After pose estimation, each frame is categorized as a common frame, mapper frame, or keyframe. 
A \emph{keyframe} is created when the number of valid correspondences with the latest keyframe falls below a threshold $\tau_{\text{k}}$, following standard SLAM practice. Keyframes are passed to the backend for pose refinement and to the mapping module for reconstruction.  
A \emph{mapper frame} is selected when the frame provides sufficient parallax for reliable multi-view reconstruction.  We compute the pixel displacement between the current frame and the latest keyframe; if the 70th percentile exceeds $\tau_{\text{m}}$, the frame is promoted to a mapper frame.
Mapper frames are first processed by the backend to compute pointmap confidence and are then used in the mapping module to initialize new 3D Gaussians.
A \emph{common frame} does not meet either the keyframe or mapper criteria and is therefore used only to refine existing scene details, without introducing new structure; its role will be further elaborated in later sections.

\subsection{Backend}
\label{section:Backend}
The backend processes keyframes from the frontend to maintain a globally consistent scene and camera trajectory. 
For each incoming keyframe, it evaluates correlations with earlier ones, builds a \emph{factor graph} over the most relevant candidates, and performs global optimization to enforce multi-view consistency. 
In addition, it estimates the confidence of keyframe and mapper pointmaps, which are later used to initialize 3D Gaussian in the mapping module.

\noindent \textbf{Loop Closure and Global Bundle Adjustment.}
Given a new keyframe, the backend first updates the factor graph by connecting it to related frames, and then performs a PnP-based global bundle adjustment (BA) to refine poses, as illustrated in Fig.~\ref{fig:slam}.(b). 
If a loop closure is detected, the current keyframe is linked to its three most relevant predecessors; otherwise, it is connected only to the latest keyframe. 
Loop detection is initially performed using the Aggregated Selective Match Kernel (ASMK). A loop is declared when a previous keyframe has an ASMK score above a threshold $\tau_{\text{loop}}$ and is at least $k_{lopp}$ keyframes apart.  
To increase robustness against weak correspondences and noisy inputs, we further leverage the 3D foundation model $\pi^3$~\citep{wang2025pi3} as our \emph{loop-detection module}. 
Specifically, the current frame and the top $N_a$ candidates from ASMK are processed by $\pi^3$ to produce pointmaps, from which we select the three most geometrically consistent keyframes based on angular error following~\citep{muraiMASt3RSLAM2024}. 
These are then connected to the factor graph, yielding more reliable loop closures and reducing drift. 
More details can be found in ~\ref{sec:append_backend}.

\noindent \textbf{Pointmap Confidence.} 
We estimate pointmap confidence using reprojection error rather than relying on the confidence values predicted by MASt3R. 
When a mapper frame or keyframe is processed, its pointmap is projected onto the $N_c$ previous keyframes with the highest ASMK scores. 
For each point, we compute the reprojection errors across the $N_c$ keyframes, average them to obtain $\bar e$, and define the confidence score as 
$C = 1$ if $\bar e \le \varepsilon_c$, and $C = \tfrac{1}{\bar e - \varepsilon_c + 1}$ otherwise, 
where $\varepsilon_c$ is a predefined threshold. This reprojection-based confidence provides a more reliable measure of geometric consistency across frames.

\begin{figure*}[t!]
\includegraphics[width=\linewidth]{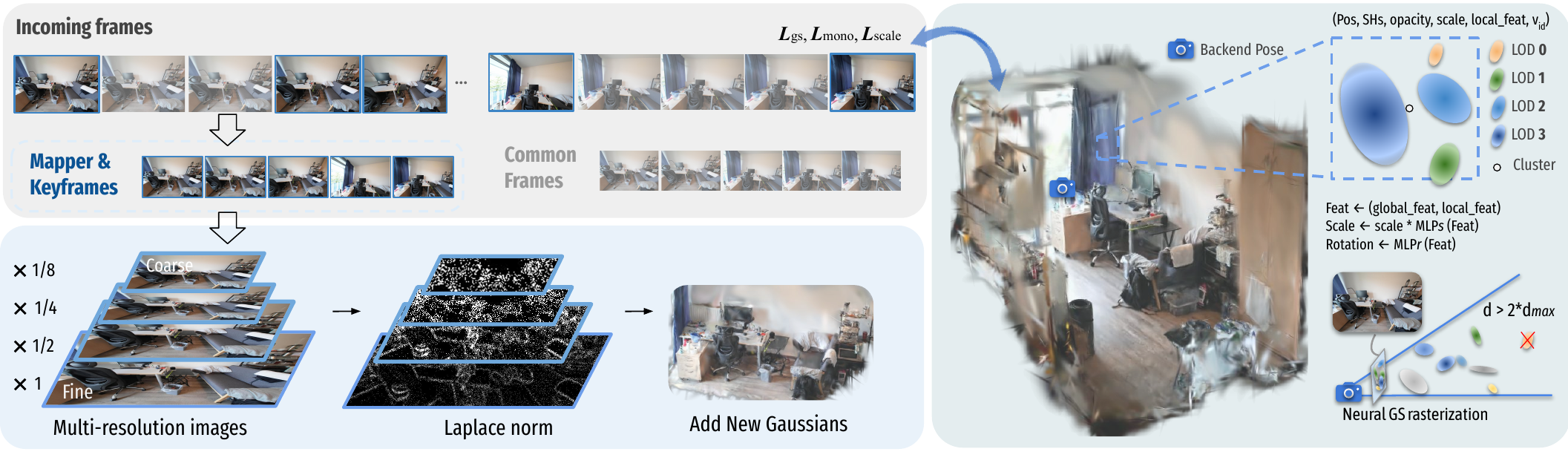}
\vspace{-12pt}
\caption{
\textbf{Mapping process.} 
When a keyframe or mapper frame arrives from the backend, new Gaussians are added to the scene. 
Multi-resolution inputs are analyzed with the Laplacian of Gaussian (LoG) operator to identify regions that require refinement, and new Gaussians are initialized at the corresponding monocular depth positions in the current view. 
Common frames are not used to add Gaussians but contribute through gradient-based refinement. 
Each primitive stores position, spherical harmonics (SH), base scale, opacity, local feature, $d_{\text{max}}$, and voxel index $v_{id}$. 
For rendering, the $d_{\text{max}}$ attribute determines whether a Gaussian is included at a given viewing distance, enabling consistent level-of-detail control.
}
\label{fig:mapping}
\vspace{-12pt}
\end{figure*}

\subsection{Mapping Module}
\label{sec:mapping}
The mapping module reconstructs the 3D Gaussian scene from incoming frames, their estimated poses, and pointmaps, as illustrated in Fig.~\ref{fig:mapping}. 
Unlike prior 3DGS-based SLAM methods that rely only on keyframes, we leverage \emph{all} frames to maximize the use of captured information: keyframes and mapper frames introduce new Gaussians, while common frames refine existing ones. 
This design enriches both visual and geometric details, which are critical not only for accurate reconstruction but also for high-fidelity rendering.  
Moreover, we introduce a hierarchical Gaussian structure with LoD-aware control. 
LoD is essential for scalable scene modeling, particularly in large-scale, navigable spaces where SLAM-based applications require consistent detail at varying viewing distances. 
By combining dense supervision from all frame types with principled level-of-detail management, our mapping module improves fidelity of both reconstructed geometry and rendered views, while maintaining computational efficiency.

\noindent \textbf{Probabilistic Selection for 3D Gaussian Insertion.} 
When a mapper frame or keyframe arrives from the backend, we determine where to initialize new 3D Gaussians. 
To avoid redundancy, Gaussians are inserted only in regions that require refinement, rather than at every pixel, guided by image-level priors inspired by~\citep{meulemanOntheflyNVS2025}. 
We prioritize high-frequency regions and poorly reconstructed areas by computing an insertion probability at each pixel $(u,v)$ using the Laplacian of Gaussian (LoG) operator~\citep{haralock1991computer}:  
\begin{equation}
P_a(u,v) = \max\!\Big(\min(\|\nabla^{2}(G_\sigma) * I(u,v)\|, 1) - \min(\|\nabla^{2}(G_\sigma) * \tilde{I}(u,v)\|, 1), 0\Big),
\end{equation}
where $I$ and $\tilde{I}$ are the ground-truth and rendered images, and $G_\sigma$ is a Gaussian kernel with standard deviation $\sigma$. 
A new Gaussian is added when $P_a(u,v)$ exceeds the threshold $\tau_{\text{a}}$.

\noindent \textbf{Gaussian Primitive Initialization.} 
After identifying candidate pixels, we initialize the corresponding 3D Gaussians. 
Each Gaussian is parameterized by its center $\mu$, spherical harmonics (SH), opacity $\alpha$, base scale $S_b$, individual feature $f_l$, and voxel index $v_{id}$. 
The $\mu$ and SH0 are initialized from the pointmap and pixel color, while opacity is set to $0.2 \cdot C_{(u,v)}$ to down-weight low-confidence regions, where $C_{(u,v)}$ is the confidence score calculated in \textit{backend}. Following~\citep{meulemanOntheflyNVS2025, wu2025monocular, yu2024mip}, the base scale at pixel $(u,v)$ is defined as:  
\begin{equation}
S_b = \frac{d_i s'}{f}, \qquad 
s' = \frac{1}{2\sqrt{\min(\|(\nabla^{2} G_\sigma) * I(u,v)\|, 1)}},
\end{equation}
where $d_i$ is the distance from the Gaussian center to the camera and $f$ is the focal length. Here, $s'$ represents an image-space scale, i.e., the expected distance to the nearest neighbor under a local 2D Poisson process of intensity $\min(\|(\nabla^{2} G_\sigma) * I(u,v)\|, 1)$,~\citep{clark1954distance}.  
To ensure smoother reconstruction, we further refine scale and initialize rotation with two MLPs:  
\begin{equation}
S = S_b \cdot \mathrm{MLP}_s(f_r \oplus f_l), \qquad 
R = \mathrm{MLP}_r(f_r \oplus f_l),
\end{equation}
where $\oplus$ denotes concatenation, $f_l$ is an individual feature initialized as zero, and $f_r$ is a region feature encoding local voxel context. 
We voxelize the 3D space with cell size $\epsilon$; when a new Gaussian is added to a voxel, the corresponding voxel-wise feature is initialized as zero and indexed by $v_{id}$. 
This hybrid region–individual design promotes global consistency of the Gaussian field while preserving local distinctiveness. 

\noindent \textbf{Levels of Detail Design.}  
To support smooth navigation in large 3D scenes, we organize Gaussians into multiple levels of detail (LoD).
Each Gaussian is assigned a level $l \in \mathbb{N}^+$ with $l < L$, where level $0$ denotes the finest resolution and level $L\!-\!1$ the coarsest. 
At initialization, a Gaussian at level $l$ corresponds to a patch of $2^{2l}$ pixels in the original image (e.g., level $0$ corresponds to one pixel). 
We progressively downsample the input frame $L-1$ times and initialize Gaussians from both the downsampled and original images.  
All Gaussian parameters follow the initialization described earlier, except that (i) the base scale is weighted by $1.4^{2l}$, and (ii) each Gaussian is assigned a distance-dependent parameter $d_{\max} = d \cdot 2^{2l}$,
where $d$ is the distance from the Gaussian center to the camera.  
During rendering, a Gaussian is included if $d_r \leq d_{\max}$, excluded if $d_r > 2d_{\max}$, and smoothly faded out for $d_{\max} < d_r \leq 2d_{\max}$ by interpolating its opacity as $\alpha = (d - d_{\max}) / d_{\max}$.  
This distance-aware LoD design suppresses flickering and maintains stable rendering quality across scales while preserving efficiency.

\noindent \textbf{Training Strategy.} 
To balance efficiency and reconstruction quality, we adopt a staged training scheme. 
For streaming input, new Gaussians are initialized and the scene is optimized for $K$ iterations whenever a mapper frame or keyframe arrives, while common frames trigger only $K/2$ iterations without inserting new Gaussians. 
Following~\citep{meulemanOntheflyNVS2025, wu2025monocular}, training frames are sampled with a 0.2 probability from the current frame and 0.8 from past frames to mitigate local overfitting. 
After processing the sequence in a streaming manner, we run a global optimization over all frames, giving higher sampling probabilities to those with fewer historical updates. 
Finally, camera poses are optimized jointly with Gaussian parameters, with gradients on positions and rotations propagated to poses, consistent with common practice in on-the-fly reconstruction.

%% file: sec/experiments.tex
\section{Experiments}
\label{sec:exp}

\subsection{Experimental Setup}

\noindent \textbf{Datasets and Metrics.} 
We evaluate on diverse indoor and outdoor benchmarks. Indoor datasets include 11 TUM scenes~\citep{sturm2012benchmark}, 14 scenes from ScanNet++~\citep{yeshwanthliu2023scannetpp}, 8 scenes from VR-NeRF~\citep{VRNeRF}, and 6 scenes from ScanNet~\citep{dai2017scannet}, with sequence lengths ranging from 32–5577 image frames. Outdoor datasets include 8 KITTI scenes~\citep{geiger2013vision}, 9 Waymo scenes~\citep{sun2020scalability} (both following S3PO-SLAM~\citep{cheng2025outdoor}), 5 scenes from Fast-livo2~\citep{zheng2024fast}, and 1 scene from MatrixCity~\citep{li2023matrixcity}, with lengths 200–1363 frames per trajectory. Reconstruction is evaluated with PSNR, SSIM~\citep{wang2004image}, and LPIPS~\citep{zhang2018unreasonable}; pose accuracy with Absolute Trajectory Error (ATE) RMSE; and system efficiency with FPS.

\noindent \textbf{Baselines.} 
We evaluate against two categories of state-of-the-art methods. For reconstruction quality, we consider 3D Gaussian Splatting approaches, including OnTheFly-NVS~\citep{meuleman2025fly}), LongSplat~\citep{lin2025longsplat}, S3PO-GS~\citep{cheng2025outdoor}, SEGS-SLAM~\citep{tianSEGSSLAM}, MonoGS~\citep{matsuki2023gaussian}). For pose estimation, in addition to the aforementioned 3DGS-based SLAM methods, we benchmark against several state-of-the-art SLAM systems, including MASt3R-SLAM~\citep{muraiMASt3RSLAM2024}, DPV-SLAM~\citep{lipson2024deep}, DROID-SLAM~\citep{teed2021droid}, and Go-SLAM~\citep{zhang2023go}.

\noindent \textbf{Implementation Details.} 
Experiments are run on a desktop with an Intel Core i9-14900K CPU and NVIDIA RTX 4090 GPU. Following standard Novel View Synthesis practices, every 8th frame is held out for evaluation: these frames are excluded from mapping but their poses are estimated and optimized for evaluation. %
In the \textit{frontend}, we set \(\tau_k = \max(0.333 \cdot W, 30)\), where \(W\) is the image width. In the \textit{backend}, we use \(N_a = \min(23, N_c)\) candidate keyframes, where \(N_c\) is the number of available candidates. If \(N_a < 11\), loop-detection modules are disabled, and the top three ASMK-scoring keyframes are directly selected to connect in the factor graph. For pointmap confidence, we fix \(\varepsilon_c = 3\).  
During \textit{mapping}, 3D Gaussians are organized into 4 LOD levels by setting \(L = 4\).

\subsection{Comparison}

\begin{figure*}[t!]
\includegraphics[width=\linewidth]{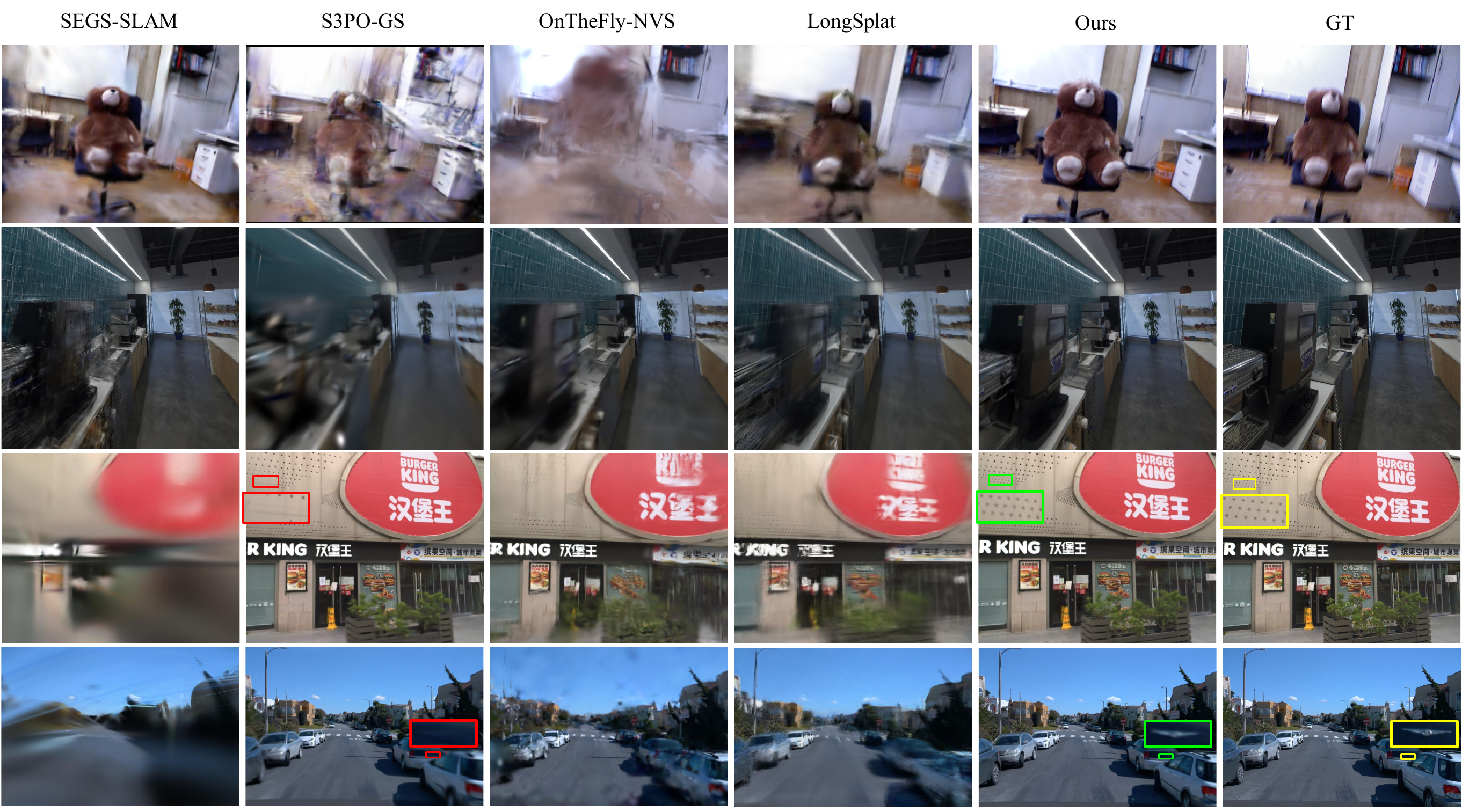}
\caption{\noindent \textbf{Qualitative comparisons} against popular on-the-fly reconstruction baselines across diverse 3D scene datasets. ARTDECO consistently preserves high-quality rendering details in complex and diverse environments, particularly in the regions highlighted with colored rectangles.
}
\label{fig:main_results}
\end{figure*}

\noindent \textbf{Reconstruction Results Analysis.} Tab.~\ref{tab:recon_all} reports reconstruction results on eight indoor and outdoor benchmarks. Our system achieves state-of-the-art quality, particularly on challenging datasets such as TUM and ScanNet with structural complexity, motion blur, and noise. On higher-quality datasets like VR-NeRF and ScanNet++, where scenes feature diverse multi-scale visuals, all methods improve, yet ARTDECO still delivers the best performance.
Outdoor evaluation covers large-scale free-motion captures (Fast-LIVO2) and forward-facing driving datasets (Waymo, KITTI, MatrixCity). ARTDECO consistently outperforms baselines, demonstrating robustness to scale variation.  
Qualitative comparisons (Fig.~\ref{fig:main_results}) show that ARTDECO, enabled by its multi-level Gaussian primitive design, captures fine details, large-scale structures, and high-fidelity geometry within a compact representation. Additional results are provided in Tabs.~\ref{tab:track_fast} - \ref{tab:track_scannetpp_lpips}, Fig.\ref{fig:qualtive_rendering_2} in ~\ref{sec:append_experiment}.
\begin{table}[t!]
\centering
\renewcommand{\arraystretch}{1.15}
\setlength{\tabcolsep}{1pt}
\caption{\textbf{Rendering comparisons against baselines} across indoor and outdoor datasets. 
We report visual quality metrics, average running time.}
\label{tab:recon_all}

\resizebox{\linewidth}{!}{
\begin{tabular}{l|ccc|ccc|ccc|ccc|c}
\toprule
Indoor-dataset & \multicolumn{3}{c|}{ScanNet++} & \multicolumn{3}{c|}{ScanNet} & \multicolumn{3}{c|}{TUM} & \multicolumn{3}{c|}{VR-NeRF} & \multicolumn{1}{c}{Training} \\
Method & PSNR$\uparrow$ & SSIM$\uparrow$ & LPIPS$\downarrow$ 
       & PSNR$\uparrow$ & SSIM$\uparrow$ & LPIPS$\downarrow$ 
       & PSNR$\uparrow$ & SSIM$\uparrow$ & LPIPS$\downarrow$ 
       & PSNR$\uparrow$ & SSIM$\uparrow$ & LPIPS$\downarrow$
       & Time$\downarrow$\\
\midrule
MonoGS            &16.71 &0.682 &0.600 &$18.87^*$ &$0.780^*$ &$0.629^*$ &17.78 &0.602 &0.573               &13.88 &0.560 &0.420             &  14.08 min \\
S3PO-GS           &\underline{22.94} &\underline{0.820} &\underline{0.355} &\underline{20.14} &\underline{0.797} &0.558             &19.62 &0.656 &0.466               &12.43 &0.642 &0.497          &  41.25 min \\
SEGS-SLAM         &-     &-     &-     &$19.73^*$ &$0.839^*$&$0.365^*$ &$19.69^*$ &$0.743^*$  &$0.307^*$  &$31.62^*$ &$0.896^*$ &$0.232^*$ & 10.84 min \\
\midrule
OnTheFly-NVS      &18.01 &0.761        &0.386 &15.36 &0.708 &\underline{0.494}             &19.72 &0.719 &0.380               &\underline{27.30} &\underline{0.872} &\underline{0.310}             &  \textbf{2.29 min} \\
LongSplat         &$24.94^*$ &$0.827^*$ &$0.260^*$               &$19.27^*$ &$0.754^*$ &$0.404^*$ &\underline{25.09} &\underline{0.804} &\underline{0.272}               &$25.74^*$ &$0.832^*$ &$0.321^*$  & 442.96 min \\
\midrule
Ours              &\textbf{29.12} &\textbf{0.918} &\textbf{0.167} &\textbf{24.10} &\textbf{0.865} &\textbf{0.271}             &\textbf{26.18} &\textbf{0.850} &\textbf{0.224}               &\textbf{28.57} &\textbf{0.895} &\textbf{0.242}             &  \underline{5.33 min}\\
\bottomrule
\end{tabular}}

\vspace{6pt} 

\resizebox{\linewidth}{!}{
\begin{tabular}{l|ccc|ccc|ccc|ccc|c}
\toprule
Outdoor-dataset & \multicolumn{3}{c|}{KITTI} & \multicolumn{3}{c|}{Waymo} & \multicolumn{3}{c|}{Fast-LIVO2} & \multicolumn{3}{c|}{MatrixCity} & \multicolumn{1}{c}{Training}\\
Method & PSNR$\uparrow$ & SSIM$\uparrow$ & LPIPS$\downarrow$ 
       & PSNR$\uparrow$ & SSIM$\uparrow$ & LPIPS$\downarrow$ 
       & PSNR$\uparrow$ & SSIM$\uparrow$ & LPIPS$\downarrow$ 
       & PSNR$\uparrow$ & SSIM$\uparrow$ & LPIPS$\downarrow$
       & Time $\downarrow$ \\
\midrule
MonoGS       &14.56 &0.489 &0.767 &19.34 &0.752 &0.627                &18.87 &0.598 &0.699                &19.36 &0.593 &0.736 &  16.52 min \\
S3PO-GS      &\underline{19.97} &\underline{0.645} &\underline{0.410} &\underline{27.28} &\underline{0.865} &0.352                &21.51 &0.684 &0.445                &21.76 &0.661 &0.584 &  34.89 min \\
SEGS-SLAM    &14.03 &0.463 &0.488 &$19.01^*$ &$0.698^*$ &$0.502^*$    &$24.58^*$  &$0.773^*$  &$0.307^*$  &\underline{25.57} &\underline{0.784} &\underline{0.366} &    8.75 min  \\
\midrule
OnTheFly-NVS &16.89 &0.579 &0.471 &25.53 &0.820 &0.360                &18.76 &0.618 &0.497                &21.36 &0.687 &0.451 & \textbf{0.74 min} \\
LongSplat    &16.86 &0.532 &0.447 &25.61 &0.795 &\underline{0.326}                &\underline{26.37} &\underline{0.792} &\underline{0.276}                &-     &-     &-     &    313.60 min  \\
\midrule
Ours         &\textbf{23.17} &\textbf{0.765} &\textbf{0.299} &\textbf{28.75} &\textbf{0.880} &\textbf{0.276}                &\textbf{29.54} &\textbf{0.894} &\textbf{0.158}                &\textbf{25.62} &\textbf{0.790} &\textbf{0.327} &  \underline{6.58 min} \\
\bottomrule
\end{tabular}}
\begin{tablenotes}
\footnotesize
\item *: majority of scenes successful; –: majority failed; Only compare fully successful methods.
\end{tablenotes}
\vspace{-6pt}
\end{table}

\noindent \textbf{Tracking Results Analysis.} Tab.~\ref{tab:track_and_memory_all} summarizes the tracking performance on indoor and outdoor benchmarks. With loop closure and covariance-matrix filtering, ARTDECO achieves markedly higher localization accuracy than other 3DGS-based systems on challenging multi-scale indoor datasets (TUM, ScanNet++). On outdoor datasets such as Waymo, it also delivers competitive performance. Further results on TUM (Second part of Tab.~\ref{tab:track_and_memory_all}) demonstrate that ARTDECO consistently outperforms state-of-the-art non-3DGS SLAM methods, confirming its superior localization capability. Per-scene metrics, additional tracking results and qualitative trajectory comparisons are provided in Tabs.~\ref{tab:track_scannetpp}--\ref{tab:track_tum_classic}, Figs.~\ref{fig:trjectory_SCANNETPP}--\ref{fig:trjectory_TUM} in \ref{sec:append_experiment}.

\begin{table}[t!]
\centering
\renewcommand{\arraystretch}{0.9}
\setlength{\tabcolsep}{6pt}
\caption{\textbf{Tracking comparisons.} 
For tracking evaluation, we compare against SLAM- and SFM-based 3D reconstruction methods on indoor and outdoor datasets, as well as state-of-the-art SLAM systems on the TUM dataset (Following MASt3R-SLAM, 9 scenes from TUM fr1). Our method consistently achieves lower ATE RMSE.}
\label{tab:track_and_memory_all}

\renewcommand{\arraystretch}{1.15}
\setlength{\tabcolsep}{6pt}
\resizebox{\linewidth}{!}{
\begin{tabular}{l|cccc|cc|c}
\toprule
Dataset & MonoGS & S3PO-GS & SEGS-SLAM & MASt3R-SLAM & OnTheFly-NVS & LongSplat & Ours \\
\midrule
ScanNet++ &1.217 &0.632 &0.245 & \underline{0.025} &0.891 &0.602 & \textbf{0.018} \\
TUM       &0.244 &0.117 &$0.073^*$ &\underline{0.031} &-     &-     & \textbf{0.025} \\
Waymo     &7.370 &\underline{1.236} &-     &-     &3.118 &4.956 & \textbf{1.213} \\
\bottomrule
\end{tabular}}

\vspace{6pt} 

\renewcommand{\arraystretch}{1.15}
\setlength{\tabcolsep}{9pt}
\resizebox{\linewidth}{!}{
\begin{tabular}{c|cccccc}
\toprule
Metric & ORB-SLAM3 & DPV-SLAM++ & DROID-SLAM & Go-SLAM & MASt3R-SLAM & Ours \\
\midrule
ATE RMSE & - & 0.054 & 0.038 & 0.035 & \underline{0.030} & \textbf{0.028} \\
\bottomrule
\end{tabular}}
\begin{tablenotes}
\footnotesize
\item *: majority of scenes successful; –: majority failed; Only compare fully successful methods.
\end{tablenotes}
\vspace{-12pt}
\end{table}

\noindent \textbf{Runtime Analysis.}
We compare runtime across 3DGS-based methods on both indoor and outdoor datasets (Tab.~\ref{tab:recon_all}). ARTDECO runs faster than all except OnTheFly-NVS, with its extra time cost primarily from pose estimation, a trade-off justified by the superior pose accuracy in Tab.~\ref{tab:track_and_memory_all}.

\subsection{Ablation Study}
\label{sec:ablation}

\begin{table}[t!]
\centering
\renewcommand{\arraystretch}{0.9}
\setlength{\tabcolsep}{6pt}
\caption{\textbf{Quantitative results on ablation studies.} We separately listed the rendering metrics and ATE RMSE on ScanNet++ dataset for each ablation described in Sec.~\ref{sec:ablation}}
\label{tab:ablation}

\renewcommand{\arraystretch}{1.15}
\setlength{\tabcolsep}{8pt}
\resizebox{\linewidth}{!}{
\begin{tabular}{l|ccccc}
\toprule
Front$\&$Backend & Full & w/ SLAM (MASt3R $\rightarrow \pi^3$) & w/ Loop ($\pi^3 \rightarrow$ vggt) & w/o loop & w/ dense key frame \\
\midrule
ATE RMSE & \textbf{0.018} & 0.374 & 0.096 & 0.057 & 0.094 \\
\bottomrule
\end{tabular}}

\vspace{6pt} 

\renewcommand{\arraystretch}{1.15}
\setlength{\tabcolsep}{6pt}
\resizebox{\linewidth}{!}{
\begin{tabular}{l|cccccc}
\toprule
Mapper & Full & w/o level-of-detail & w/o implicit structure & w/o global feat & w/o mapper frame & w/o common frame \\
\midrule
PSNR & \textbf{29.12} & 28.13 & 28.54 & 27.95 & 26.38 & 27.20  \\
SSIM & \textbf{0.918} & 0.912 & 0.914 & 0.910 & 0.898 & 0.904 \\
LPIPS & \textbf{0.167} & 0.180 & 0.175 & 0.197 & 0.229 & 0.211 \\
\bottomrule
\end{tabular}}
\vspace{-12pt}

\end{table}

\noindent \textbf{Ablation on Localization.}
We analyze the impact of backbone choice, loop closure, and frame categorization strategy on localization, as summarized in Tab.~\ref{tab:ablation}. We first ablate the feed-forward model used in the \textit{frontend} and \textit{backend} by replacing MASt3R (pairwise inference) with $\pi^{3}$ (multi-image inference). Although $\pi^{3}$ is trained on more diverse data, it lacks metric-scale capability and performs worse under varying viewpoints. In contrast, MASt3R better preserves consistent object proportions, resulting in more accurate pose estimation. Next, we ablate the loop-closure module by disabling it, which leads to a significant degradation in localization accuracy. Finally, we ablate the frame categorization strategy. Here, MF denotes mapper frames and KF denotes keyframes. Using both MFs and KFs (track w/ MF\&KF) for inference provides additional temporal information, but unexpectedly reduces pose accuracy. This is because 3D foundation models often struggle with small-parallax inputs, producing ghosting and blur that corrupt point clouds and feature correspondences when the input sequence is overly dense.

\noindent \textbf{Ablation on Reconstruction.}
We further ablate the effects of mapper frames, level-of-detail, and structural Gaussians on reconstruction, as shown in Tab.~\ref{tab:ablation}. MFs add richer multi-view constraints, while LoD and structural Gaussians yield more compact, regularized representations, together improving reconstruction fidelity and rendering quality.

%% file: sec/limitation.tex
\section{Limitations}
\label{sec:limitation}
While ARTDECO achieves strong reconstruction and localization, it has several limitations. First, it partly depends on feed-forward 3D foundation models for correspondence and geometry, which, despite enabling fast and scalable inference, reduce robustness under noise, blur, or lighting changes, and suffer when inputs fall outside the training distribution. Second, the system assumes consistent illumination and sufficient parallax; violations such as low-texture surfaces, repetitive structures, or near-degenerate trajectories can cause drift or artifacts. These challenges suggest future work on incorporating uncertainty estimation, adaptive model selection, and stronger priors to improve generalization and reliability in real-world settings.

%% file: sec/conclusion.tex
\section{Conclusion}
\label{sec:conclusion}
In this work, we present ARTDECO, a unified framework that advances on-the-fly 3D reconstruction from monocular image sequences. Beyond achieving strong results on standard indoor and outdoor benchmarks, ARTDECO demonstrates that feed-forward priors and structured Gaussian representations can be effectively combined within a single system to deliver both accuracy and efficiency. 
We see ARTDECO as a step toward practical large-scale deployment of real-to-sim pipelines, with promising applications in AR/VR, robotics, and digital twins.

%% file: sec/supplementary.tex
\section{Supplementary Material}

We organize the supplementary material as follows:
\begin{adjustwidth}{0.2em}{0pt}
\begin{itemize}
    \item \textbf{Sec.~\ref{sec:implementation}} describes implementation \& evaluation protocol.
    \item \textbf{Sec.~\ref{sec:jacobians}} derives Jacobians and covariance transformation used in the Frontend / Backend.
    \item \textbf{Sec.~\ref{sec:pi3}} presents additional experimental details, including $\pi^3$-based multi-frame ablations and their observations.
    \item \textbf{Sec.~\ref{sec:append_frontend}} expands the Frontend implementation, covering correspondence formation, residual weighting, Gauss-Newton updates, and focal optimization under unknown intrinsics.
    \item \textbf{Sec.~\ref{sec:append_backend}}describes Backend loop-closure detection (ASMK + multi-frame 3D priors) and global bundle adjustment.
    \item \textbf{Sec.~\ref{sec:append_experiment}} reports extended experiments, per-scene quantitative metrics, and additional qualitative results.
\end{itemize}
\end{adjustwidth}

\subsection{More Implementation Details}
\label{sec:implementation}


\paragraph{Evaluation protocol.}
Prior works evaluate reconstruction quality on different subsets of frames, as each method selects its own keyframes for mapping. This inconsistency leads to slight but systematic metric bias. To ensure a fair and reproducible evaluation, we uniformly sample one frame out of every eight across all sequences as evaluation frames. These frames are excluded from mapping supervision and are used only for pose optimization during inference.

We have re-implemented or modified baseline evaluation scripts accordingly so that all reported metrics in Tables are computed under this identical protocol. This guarantees that differences in performance arise solely from algorithmic design rather than evaluation selection bias.

\subsection{Jacobian}
\label{sec:jacobians}

Let a 3D point in the current camera coordinate system be
$\mathbf{x} = (X, Y, Z)^{\top}$.
Its projection to the image plane with log–depth parameterization is defined as
\begin{equation}
\pi(X,Y,Z) =
\begin{bmatrix}
f_x X / Z + c_x \\[2pt]
f_y Y / Z + c_y \\[2pt]
\log Z
\end{bmatrix},
\end{equation}
where $(f_x,f_y)$ are focal lengths and $(c_x,c_y)$ are principal point offsets.  
The differential of this mapping governs how small 3D perturbations propagate into pixel-space.

\paragraph{Jacobian with respect to predicted 3D points.}
The Jacobian of the projection function w.r.t.\ the 3D coordinates is
\begin{equation}
J_{\pi}(\mathbf{x}) =
\begin{bmatrix}
\frac{f_x}{Z} & 0 & -\frac{f_x X}{Z^2} \\[4pt]
0 & \frac{f_y}{Z} & -\frac{f_y Y}{Z^2} \\[4pt]
0 & 0 & \frac{1}{Z}
\end{bmatrix}.
\label{eq:jacobian_points}
\end{equation}

Given the per-point covariance $\Sigma_c \in \mathbb{R}^{3\times3}$ predicted in the current frame
(estimated from local neighborhoods as described in Sec.~3.1), the measurement-space covariance
after projection to the keyframe view $k$ is
\begin{equation}
\Sigma_{ck} = J_{\pi}(\mathbf{x}_{ck})\, R_{kc}\, \Sigma_c \, R_{kc}^{\top} \, J_{\pi}(\mathbf{x}_{ck})^{\top},
\end{equation}
where $R_{kc}$ is the rotation matrix of the relative pose $T_{kc}$ between the current frame
$c$ and keyframe $k$.
We reject a correspondence if $\det(\Sigma_{ck}) > \tau$, which effectively discards
measurements with high geometric uncertainty while remaining rotation-invariant.


\paragraph{Jacobian with respect to focal length}
When the focal length is unknown or weakly calibrated, we jointly optimize it with the pose
parameters.  
Let a pixel $\mathbf{p}_c=(u_c,v_c)$ in the current frame correspond to the 3D point
$\mathbf{P}_c=(X_c,Y_c,Z_c)^{\top}$.
The point reconstructed in normalized camera coordinates is
\begin{equation}
\mathbf{P}'_c = Z_c
\begin{bmatrix}
(u_c - c_x)/f \\[2pt]
(v_c - c_y)/f \\[2pt]
1
\end{bmatrix},
\end{equation}
and its transformation to the keyframe $k$ is
$\mathbf{P}_k = T_{kc}\,\mathbf{P}'_c$ with
$T_{kc} = [R_{kc},\,t_{kc}] \in \mathrm{Sim}(3)$.
The projection back to pixel space is
\begin{equation}
\pi(\mathbf{P}_k) =
\begin{bmatrix}
f X_k/Z_k + c_x \\[2pt]
f Y_k/Z_k + c_y \\[2pt]
\log Z_k
\end{bmatrix}.
\end{equation}





The focal length $f$ influences the projection in two ways: (1) \textbf{Direct effect:} scaling of projected pixel coordinates. (2) \textbf{Indirect effect:} change in the reconstructed 3D point
$\mathbf{P}'_c$ itself, which depends on $f$ through normalization.

The corresponding partial derivatives are
\begin{align}
\frac{\partial (u,v,\log Z)}{\partial f}
&=
\begin{bmatrix}
X_k/Z_k \\[2pt] Y_k/Z_k \\[2pt] 0
\end{bmatrix},
\\[4pt]
\frac{\partial \mathbf{P}'_c}{\partial f}
&= Z_c
\begin{bmatrix}
-\tfrac{(u_c - c_x)}{f^2} \\[2pt]
-\tfrac{(v_c - c_y)}{f^2} \\[2pt]
0
\end{bmatrix},
\quad
\frac{\partial \mathbf{P}_k}{\partial f}
= T_{kc}\, \frac{\partial \mathbf{P}'_c}{\partial f}.
\end{align}

Combining both contributions via the chain rule yields the full Jacobian:
\begin{equation}
J_{\pi,f}(\mathbf{P}_k)
=
\underbrace{
\frac{\partial (u,v,\log Z)}{\partial f}
}_{\text{direct}}
+
\underbrace{
\frac{\partial (u,v,\log Z)}{\partial \mathbf{P}_k}
\frac{\partial \mathbf{P}_k}{\partial f}
}_{\text{indirect}}
=
\begin{bmatrix}
\frac{X^k}{Z^k} \\[6pt]
\frac{Y^k}{Z^k} \\[6pt]
0
\end{bmatrix}
+
\begin{bmatrix}
\frac{f}{Z^k} & 0 & -\frac{f X^k}{(Z^k)^2} \\[6pt]
0 & \frac{f}{Z^k} & -\frac{f Y^k}{(Z^k)^2} \\[6pt]
0 & 0 & \frac{1}{Z^k}
\end{bmatrix}
\begin{bmatrix}
-\tfrac{(u^c - c_x)}{f^2} \\[6pt]
-\tfrac{(v^c - c_y)}{f^2} \\[6pt]
0
\end{bmatrix}
\label{eq:jacobian_focal}
\end{equation}
This joint derivative allows the optimizer to update focal length coherently with the
camera pose.

\subsection{Additional Experiments}
\label{sec:pi3}

Sec.~\ref{sec:ablation} of the main paper evaluates how different architectural and procedural choices
affect localization and reconstruction.  
Here we provide detailed background and rationale for one
particularly important ablation: replacing the pairwise correspondence model MASt3R
with the multi-frame visual-geometry model $\pi^3$.  
The goal is to understand whether stronger 3D priors and longer temporal context
can further enhance robustness and global consistency in the pipeline.

\paragraph{Background.}
MASt3R~\citep{mast3r_eccv24} predicts two-view correspondences and metric pointmaps
with well-calibrated scale but limited temporal context.
In contrast, $\pi^3$~\citep{wang2025pi3} is a permutation-equivariant large-scale
geometry model trained on multi-image sets.
Its potential advantage lies in leveraging more global spatial constraints and recovering
denser, smoother pointmaps across frames.
However, because $\pi^3$ is not explicitly metric-aware, integrating it into a streaming
SLAM-style system introduces challenges in maintaining scale and temporal consistency.
Below we describe how we adapt $\pi^3$ to our framework
and analyze its effect relative to the baseline.


\paragraph{Frontend with $\pi^3$ Inference and Keyframe Scheduling.}
In the original frontend, each incoming frame is matched
against the latest keyframe using MASt3R.
To enable multi-frame inference, we accumulate $k$ consecutive frames before calling $\pi^3$.
The model jointly estimates pointmaps for all $k$ views and provides dense pixel correspondences.
After inference, we perform a local bundle adjustment (LBA) in the $\mathrm{Sim}(3)$
group among these $k$ frames. Upon completing this process, we follow the keyframe selection strategy described in Sec.~\ref{section:fronted}, and the selected keyframes are then passed to the backend. 

After the first $\pi^3$ inference, subsequent updates reuse the $l$ most recent keyframes
as temporal anchors and wait until $k{-}l$ new frames arrive before triggering the next
inference batch.
This overlap ensures continuity and smooth transition of geometric priors between windows.



\paragraph{Backend with $\pi^3$-Based Global Optimization.}
The backend receives keyframes streamed from the frontend.
To incorporate $\pi^3$ priors, it similarly waits until $k$ keyframes are accumulated,
then performs a joint $\pi^3$ inference to obtain multi-view correspondences among
the current and historical keyframes.
A global bundle adjustment (GBA) then optimizes all selected poses.
Compared with the frontend’s local window, the backend’s inference window integrates non-adjacent keyframes filtered by the ASMK module from MASt3R-SLAM~\citep{muraiMASt3RSLAM2024}, providing long-range loop-closure evidence while keeping runtime tractable.



\paragraph{Mapper with Keyframe Reception.}  
The mapping module remains unchanged.
The mapper simply receives the keyframes sent by the backend and proceeds with the reconstruction following the process described in Sec.~\ref{sec:mapping}.

\subsection{Details in Frontend:  Pose and Focal Optimization}
\label{sec:append_frontend}


The frontend provides the first stage of the streaming pipeline.
We now present a detailed formulation of the weighted Gauss–Newton optimization
that simultaneously refines pose and focal length when intrinsics are unknown.

For each input frame, MASt3R predicts dense two-view correspondences
and associated 3D points between the current frame \(c\) and the latest keyframe \(k\).
We denote the correspondence set as
\begin{equation}
    \mathcal{C}=\{(\mathbf{p}_c^m,\mathbf{p}_k^m,
\mathbf{P}_c^m,\mathbf{P}_k^m)\}_{m=1}^{M},
\end{equation}

where \(\mathbf{p}_c^m,\mathbf{p}_k^m\in\mathbb{R}^2\) are pixel coordinates and
\(\mathbf{P}_c^m,\mathbf{P}_k^m\in\mathbb{R}^3\) are the corresponding 3D points
expressed in the coordinate systems of the current frame and keyframe, respectively.
The relative transform \(T_{kc}=(s_{kc},\mathbf{R}_{kc},\mathbf{t}_{kc})\)
includes scale, rotation, and translation components.

\paragraph{Residual formulation.}
For each correspondence \(m\),
we compose the similarity transform and projection directly into the residual:
\begin{equation}
\mathbf{r}_m=
\begin{bmatrix}
\mathbf{p}_k^m-\hat{\mathbf{p}}_k^m\\[2pt]
\log Z_k^m-\log\hat{Z}_k^m
\end{bmatrix},
\qquad
\hat{\mathbf{P}}_k^m=s_{kc}\mathbf{R}_{kc}\mathbf{P}_c^m+\mathbf{t}_{kc},
\qquad
\hat{\mathbf{p}}_k^m=\pi_{K(f)}(\hat{\mathbf{P}}_k^m),
\label{eq:a4_residual}
\end{equation}
where \(\pi_{K(f)}(\cdot)\) denotes the pinhole projection with intrinsic
matrix \(K(f)\).
The objective seeks the optimal parameters that minimize
the robust weighted energy
\begin{equation}
E_{\mathrm{rob}}=\tfrac{1}{2}\sum_{m=1}^{M}
\mathbf{r}_m^{\!\top}\mathbf{W}_m^{\mathrm{rob}}\mathbf{r}_m,
\qquad
\mathbf{W}_m^{\mathrm{rob}}=\omega_m\mathbf{W}_m .
\label{eq:a4_energy}
\end{equation}

\paragraph{Weighting and robustness.}
Each correspondence is assigned a positive-semidefinite weight matrix
\(\mathbf{W}_m\in\mathbb{R}^{3\times3}\) derived from measurement covariance,
balancing pixel and log-depth residuals.
To suppress outliers, we apply a Huber kernel:
\begin{equation}
    \omega_m=
\begin{cases}
1, & s_m\le\delta,\\[2pt]
\dfrac{\delta}{s_m+\varepsilon}, & s_m>\delta,
\end{cases}
\qquad
s_m=\sqrt{\mathbf{r}_m^{\!\top}\mathbf{W}_m\mathbf{r}_m},
\end{equation}

with a small \(\varepsilon>0\) for numerical stability.

\paragraph{Linearization and update.}
Linearizing Eq.~\ref{eq:a4_residual} about the current estimate yields
the normal equations
\begin{equation}
\Big(\sum_m\mathbf{J}_m^{\!\top}\mathbf{W}_m^{\mathrm{rob}}\mathbf{J}_m\Big)\Delta\boldsymbol{\theta}
=\sum_m\mathbf{J}_m^{\!\top}\mathbf{W}_m^{\mathrm{rob}}\mathbf{r}_m,
\label{eq:a4_normal}
\end{equation}
where \(\mathbf{J}_m=\partial\mathbf{r}_m/\partial\boldsymbol{\theta}\)
and \(\boldsymbol{\theta}=\{T_{kc},f\}\).
For pose-only optimization, the dimensionality is \(d=7\);
when focal refinement is enabled, \(d=8\).
The updates are applied through the exponential map in the
\(\mathrm{Sim}(3)\) Lie group:
\begin{equation}
    T_{kc}\leftarrow\exp_{\mathrm{Sim(3)}}(\Delta\boldsymbol{\xi}_{\mathrm{sim}})\,T_{kc},
\qquad
f\leftarrow f+\Delta f,
\end{equation}

where \(\Delta\boldsymbol{\xi}_{\mathrm{sim}}\in\mathbb{R}^{7}\) is the minimal increment of the similarity transform.

\subsection{Details in Backend: Loop Closure and Global Bundle Adjustment}
\label{sec:append_backend}

The long-term trajectory accuracy requires closing loops and enforcing multi-view
consistency across the entire sequence.
The backend of ARTDECO detects loop closures, verifies them
with 3D priors, and performs global optimization of all keyframe poses
in the $\mathrm{Sim}(3)$ group.

\paragraph{Hybrid Loop-Closure Detection.}
Given a new keyframe $K_t$, the backend first computes its ASMK similarity
to all historical keyframes $\{K_j\}$.
Candidates with a score greater than $0.005$ are retained.
If the temporal gap between $K_t$ and the nearest candidate exceeds 10 frames,
we assume a potential loop and re-rank the top $N_a$ candidates
by similarity.

Next, we perform $\pi^3$ inference jointly on the set
$\{K_t\}\cup\mathcal{C}_{N_a}$ to obtain dense pointmaps in a shared
coordinate system.
For each candidate keyframe $K_j$, we compute the angular and depth errors
between corresponding 3D points $\mathbf{P}_t^m$ and $\mathbf{P}_j^m$.
If the ratio of geometrically consistent correspondences exceeds $0.15$,
the pair $(K_t,K_j)$ is confirmed as a loop closure and added as an edge
in the factor graph.

This two-stage scheme achieves a balance between robustness and efficiency:
ASMK rapidly filters potential loops,
and $\pi^3$ provides dense geometric validation resilient in practice.

\paragraph{Global Bundle Adjustment.}
After loop-closure edges are added,
we jointly refine all connected keyframe poses
$\{T_{wi}\}=\{(s_i,\mathbf{R}_i,\mathbf{t}_i)\}\in\mathrm{Sim}(3)$
through global bundle adjustment (GBA).
For a correspondence $m$ between frames $i$ and $j$,
let $\mathbf{P}_i^m\in\mathbb{R}^3$ be the 3D point in frame~$i$
and $\mathbf{p}_j^m\in\mathbb{R}^2$ the observed pixel in frame~$j$.
The transformed and projected quantities are
\begin{equation}
\hat{\mathbf{P}}_w^m = s_i\mathbf{R}_i\mathbf{P}_i^m+\mathbf{t}_i,
\qquad
\hat{\mathbf{P}}_j^m = s_j^{-1}\mathbf{R}_j^{\!\top}
(\hat{\mathbf{P}}_w^m-\mathbf{t}_j),
\qquad
\hat{\mathbf{p}}_j^m = \pi_K(\hat{\mathbf{P}}_j^m).
\end{equation}
We define the residual vector as
\begin{equation}
\mathbf{r}_m =
\begin{bmatrix}
\hat{\mathbf{p}}_j^m-\mathbf{p}_j^m\\[2pt]
\log \hat{Z}_j^m-\log Z_j^m
\end{bmatrix},
\label{eq:a5_residual}
\end{equation}
which combines pixel reprojection and log-depth errors.
The objective function sums all residuals weighted by their
confidence~$\omega_m$:
\begin{equation}
E_{\mathrm{GBA}}=
\sum_{i,j,m}\omega_m\,
\mathbf{r}_m^{\!\top}\mathbf{r}_m.
\label{eq:a5_objective}
\end{equation}

For each correspondence, the projection Jacobian with respect to
the 3D point $\hat{\mathbf{P}}_j^m=(X_j,Y_j,Z_j)^{\top}$ is
\begin{equation}
\frac{\partial \mathbf{r}_m}{\partial \hat{\mathbf{P}}_j^m} =
\begin{bmatrix}
\frac{f_x}{Z_j} & 0 & -\frac{f_x X_j}{Z_j^2}\\[2pt]
0 & \frac{f_y}{Z_j} & -\frac{f_y Y_j}{Z_j^2}\\[2pt]
0 & 0 & \frac{1}{Z_j}
\end{bmatrix}.
\end{equation}
Perturbing poses in $\mathrm{Sim}(3)$ by left-multiplicative increments
$\delta\boldsymbol{\xi}_i,\delta\boldsymbol{\xi}_j\!\in\!\mathbb{R}^7$
yields
\begin{align}
\frac{\partial \hat{\mathbf{P}}_j^m}{\partial \delta\boldsymbol{\xi}_i}
&= s_j^{-1}\mathbf{R}_j^{\!\top}
\!\begin{bmatrix}
\mathbf{I}_3 & -[\hat{\mathbf{P}}_w^m]_{\times}\\[2pt]
\mathbf{0}^{\top} & \mathbf{P}_w^m
\end{bmatrix},
&
\frac{\partial \hat{\mathbf{P}}_j^m}{\partial \delta\boldsymbol{\xi}_j}
&=
\begin{bmatrix}
-\mathbf{I}_3 & [\hat{\mathbf{P}}_j^m]_{\times}
\end{bmatrix},
\end{align}
where $[\cdot]_{\times}$ denotes the skew-symmetric matrix.
By the chain rule,
\begin{equation}
    \frac{\partial \mathbf{r}_m}{\partial \delta\boldsymbol{\xi}_i}
=\frac{\partial \mathbf{r}_m}{\partial \hat{\mathbf{P}}_j^m}
\frac{\partial \hat{\mathbf{P}}_j^m}{\partial \delta\boldsymbol{\xi}_i},
\qquad
\frac{\partial \mathbf{r}_m}{\partial \delta\boldsymbol{\xi}_j}
=\frac{\partial \mathbf{r}_m}{\partial \hat{\mathbf{P}}_j^m}
\frac{\partial \hat{\mathbf{P}}_j^m}{\partial \delta\boldsymbol{\xi}_j}.
\end{equation}

After global optimization,
each keyframe~$k$ reprojects its 3D points $\{\mathbf{X}_k^n\}_{n=1}^{N_k}$
to all connected keyframes~$\mathcal{N}_k$ and computes the mean
reprojection error
\begin{equation}
e_n=\frac{1}{|\mathcal{N}_k|}
\sum_{j\in\mathcal{N}_k}
\|\mathbf{u}_j^n-\pi_K(T_{jk}\mathbf{X}_k^n)\|_2.
\end{equation}
A confidence score is then assigned as
\begin{equation}
    c_n=
\begin{cases}
1, & e_n\le\tau_{\mathrm{proj}},\\[2pt]
\dfrac{1}{e_n-\tau_{\mathrm{proj}}+1}, & e_n>\tau_{\mathrm{proj}},
\end{cases}
\end{equation}

and the weighted pairs
$\{(\mathbf{X}_k^n,c_n)\}$ are sent to the mapping thread.
This ensures that subsequent Gaussian-primitive updates favor
high-confidence geometric regions while down-weighting uncertain areas.

\subsection{More Experiments}
\label{sec:append_experiment}

Below, We provide per-scene quantitative results and qualitative comparisons to further support the main paper. As shown in Tab.~\ref{tab:track_fast}-~\ref{tab:track_tum_classic}, \textsc{ARTDECO} consistently achieves the highest PSNR and SSIM and the lowest LPIPS across all datasets, validating its strong generalization from vatious type of indoor scenes (TUM~\citep{sturm2012benchmark}, ScanNet~\citep{dai2017scannet}, ScanNet++~\citep{yeshwanthliu2023scannetpp}, VR-NeRF~\citep{VRNeRF}) to large-scale outdoor captures (Waymo~\citep{sun2020scalability}, Fast-LIVO2~\citep{zheng2024fast}, KITTI~\citep{geiger2013vision}, MatrixCity~\citep{li2023matrixcity}). 

\begin{figure*}[t!]
\includegraphics[width=\linewidth]{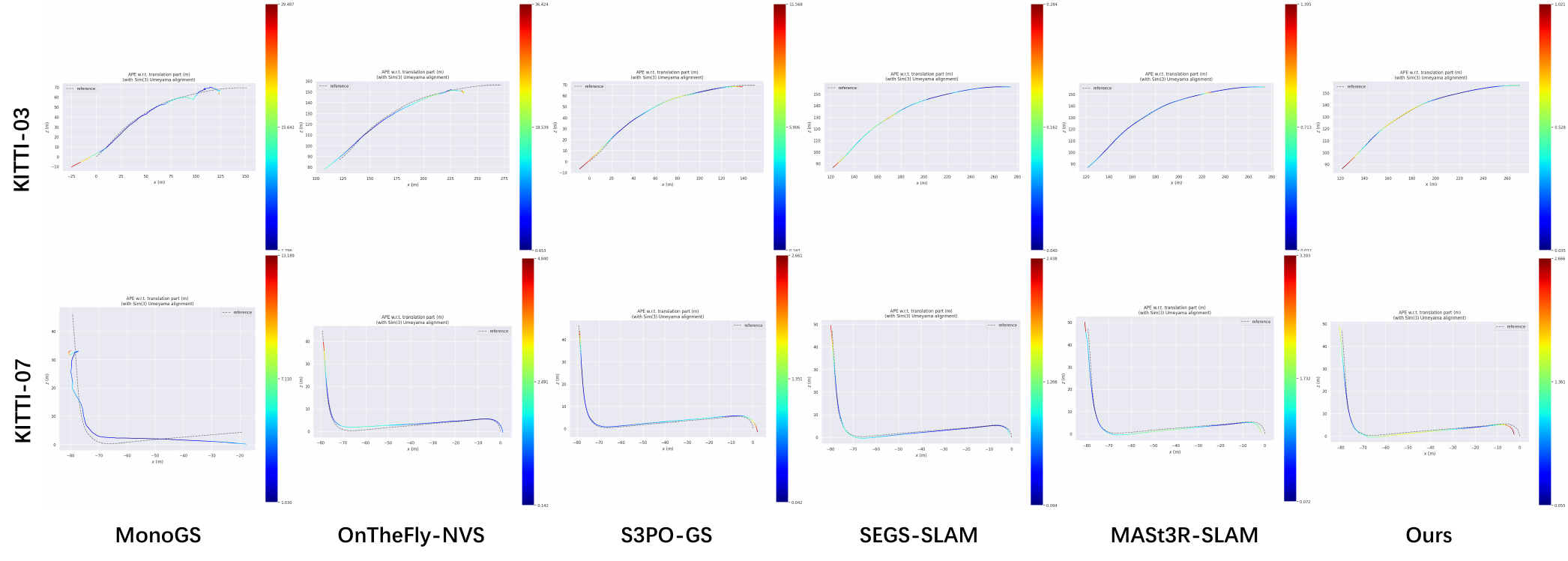}
\caption{Qualitative Comparison of Trajectories across Different Methods on the KITTI Dataset.}
\label{fig:trjectory_KITTI}
\end{figure*}

\begin{figure*}[t!]
\includegraphics[width=\linewidth]{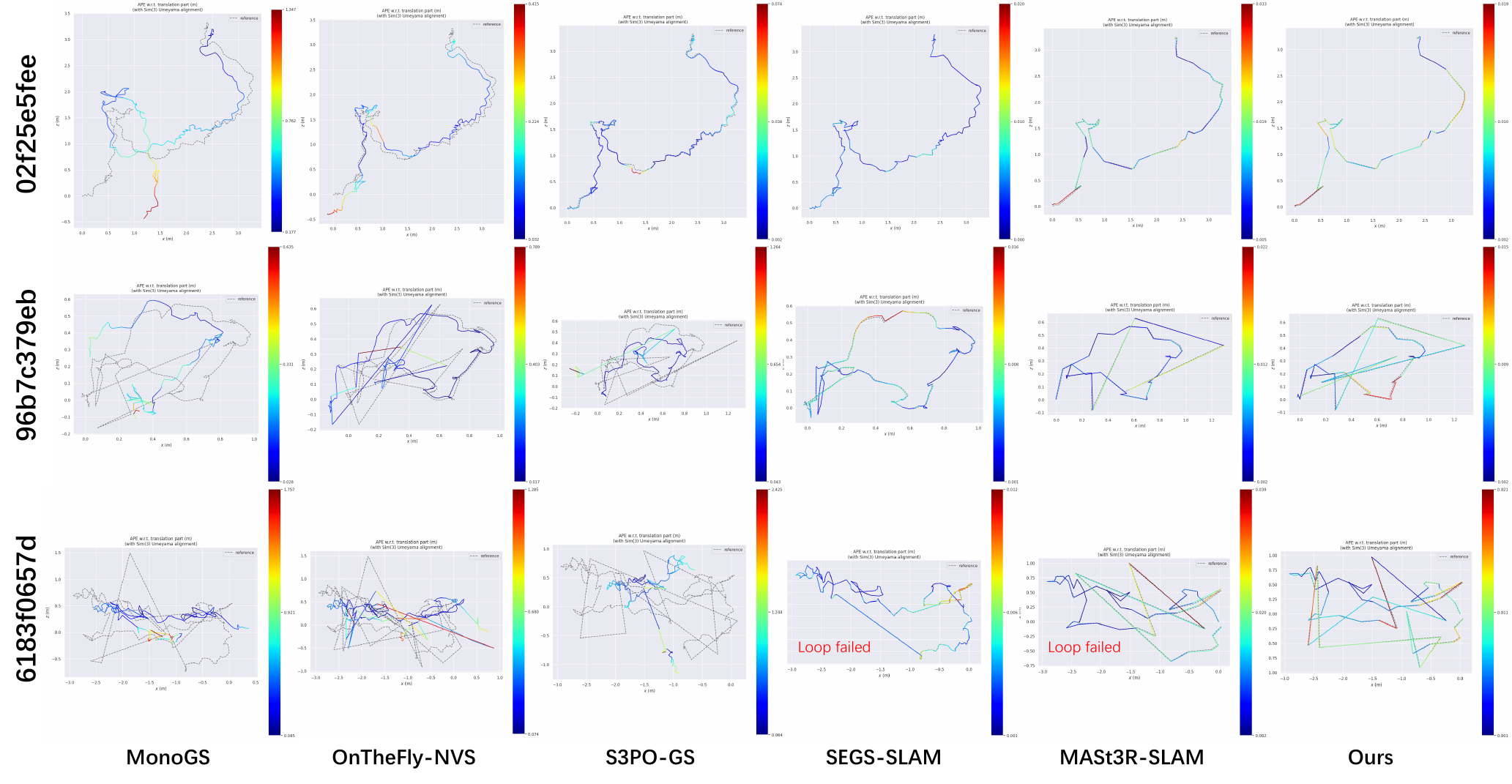}
\caption{Qualitative Comparison of Trajectories across Different Methods on the ScanNet++ Dataset.}
\label{fig:trjectory_SCANNETPP}
\end{figure*}

\begin{figure*}[t!]
\includegraphics[width=\linewidth]{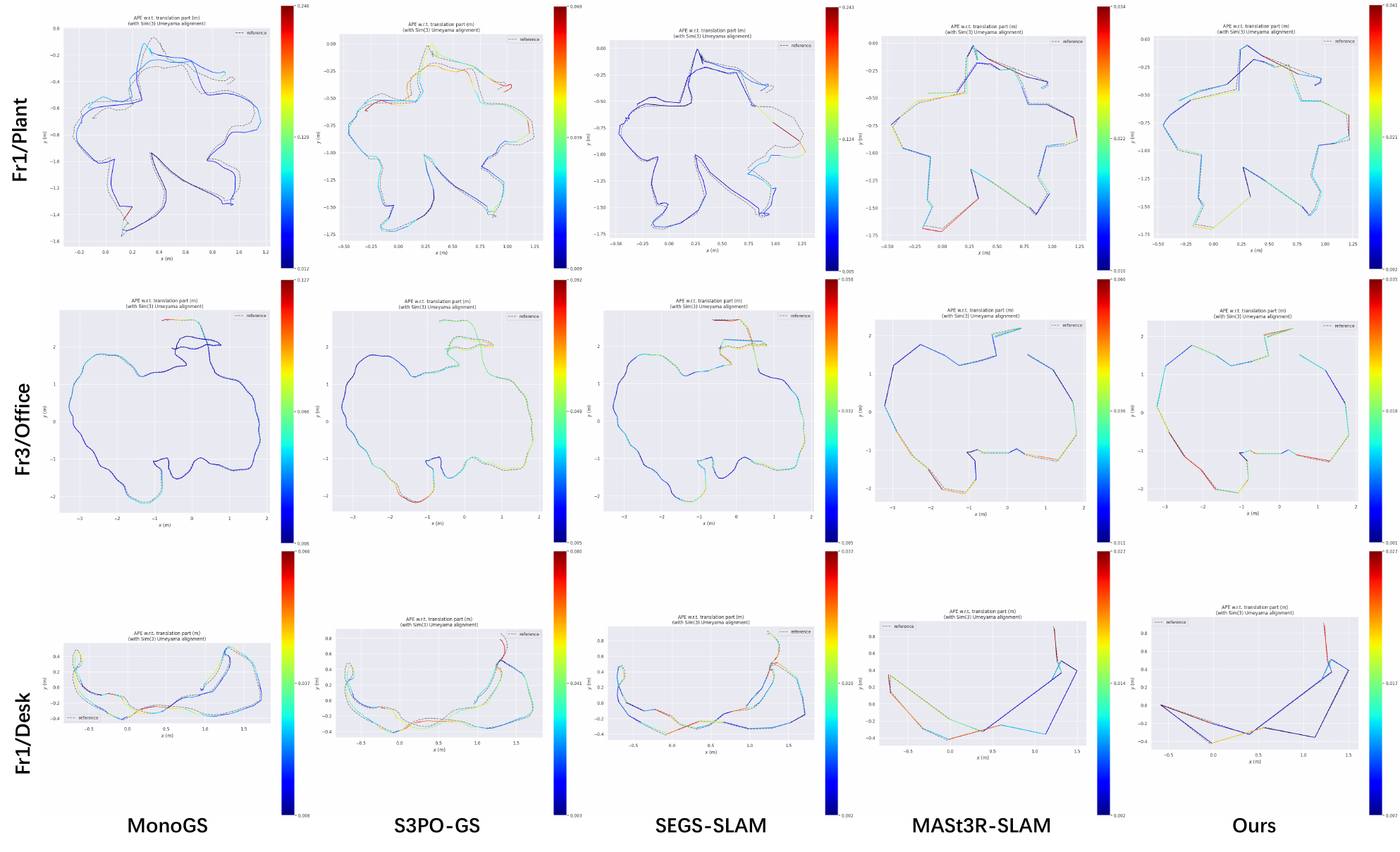}
\caption{Qualitative Comparison of Trajectories across Different Methods on the TUM Dataset.}
\label{fig:trjectory_TUM}
\end{figure*}

We also provide qualitative results related to reconstruction and tracking trajectories, as shown in Fig.~\ref{fig:trjectory_SCANNETPP}~\ref{fig:trjectory_KITTI}~\ref{fig:trjectory_TUM}.

\begin{table}[htbp]
\begin{center}
\renewcommand{\arraystretch}{1.15}
\setlength{\tabcolsep}{10pt}
\footnotesize
\caption{PSNR on the Fast-LIVO2 Dataset}   
\label{tab:track_fast}%
\resizebox{1\linewidth}{!}{
\begin{tabular}{c|ccccc}
\toprule
   Method   & CBD\_Building\_01 & HKU\_Campus & Red\_Sculpture & Retail\_Street & SYSU \\
\midrule
MonoGS & 19.86 & 21.70 & 15.50 & 18.05 & 19.23 \\

SEGS-SLAM & 26.26 & \underline{29.55} & - & 18.49 & 24.01 \\

S3PO-GS & 17.47 & 25.47 & 18.89 & \underline{24.30} & 21.42 \\

\midrule

OnTheFly-NVS & 17.79 & 21.46 & 17.50 & 17.67 & 19.39 \\

LongSplat & \underline{29.25} & 29.45 & \underline{24.97} & 23.10 & \underline{25.07} \\

\midrule

Ours & \textbf{31.11} & \textbf{30.89} & \textbf{26.29} & \textbf{29.20} & \textbf{30.22} \\
\bottomrule
\end{tabular}%
}
\end{center}
\end{table}

\begin{table}[htbp]
\begin{center}
\renewcommand{\arraystretch}{1.15}
\setlength{\tabcolsep}{10pt}
\footnotesize
\caption{SSIM on the Fast-LIVO2 Dataset}   
\label{tab:ssim_fast}%
\resizebox{1\linewidth}{!}{
\begin{tabular}{c|ccccc}
\toprule
   Method   & CBD\_Building\_01 & HKU\_Campus & Red\_Sculpture & Retail\_Street & SYSU \\
\midrule
MonoGS & 0.698 & 0.608 & 0.518 & 0.554 & 0.610 \\

SEGS-SLAM & 0.880 & \underline{0.847} & - & 0.589 & \underline{0.777} \\

S3PO-GS & 0.645 & 0.705 & 0.620 & \underline{0.782} & 0.669 \\

\midrule

OnTheFly-NVS & 0.677 & 0.628 & 0.590 & 0.558 & 0.635 \\

LongSplat & \underline{0.891} & 0.832 & \underline{0.786} & 0.708 & 0.742 \\

\midrule

Ours & \textbf{0.940} & \textbf{0.871} & \textbf{0.861} & \textbf{0.901} & \textbf{0.899} \\
\bottomrule
\end{tabular}%
}
\end{center}
\end{table}

\begin{table}[htbp]
\begin{center}
\renewcommand{\arraystretch}{1.15}
\setlength{\tabcolsep}{10pt}
\footnotesize
\caption{LPIPS on the Fast-LIVO2 Dataset}   
\label{tab:lpips_fast}%
\resizebox{1\linewidth}{!}{
\begin{tabular}{c|ccccc}
\toprule
   Method   & CBD\_Building\_01 & HKU\_Campus & Red\_Sculpture & Retail\_Street & SYSU \\
\midrule
MonoGS & 0.623 & 0.687 & 0.778 & 0.750 & 0.658 \\

SEGS-SLAM & 0.213 & \underline{0.205} & - & 0.516 & \underline{0.292} \\

S3PO-GS & 0.592 & 0.448 & 0.505 & \underline{0.232} & 0.451 \\

\midrule

OnTheFly-NVS & 0.490 & 0.470 & 0.500 & 0.528 & 0.498 \\

LongSplat & \underline{0.179} & 0.258 & \underline{0.304} & 0.315 & 0.322 \\

\midrule

Ours & \textbf{0.108} & \textbf{0.199} & \textbf{0.205} & \textbf{0.127} & \textbf{0.151} \\
\bottomrule
\end{tabular}%
}
\end{center}
\end{table}

\begin{table}[htbp]
\begin{center}
\renewcommand{\arraystretch}{1.15}
\setlength{\tabcolsep}{2pt}
\footnotesize
\caption{PSNR on the TUM Dataset}   
\label{tab:psnr_tum}%
\resizebox{1\linewidth}{!}{
\begin{tabular}{c|ccccccccccc}
\toprule
Method & f1\_360 & f1\_desk & f1\_desk2 & f1\_floor & f1\_plant & f1\_room & f1\_rpy & f1\_teddy & f1\_xyz & f2\_xyz & f3\_office \\
\midrule
MonoGS & 16.17 & 14.86 & 14.96 & 20.71 & 17.46 & 15.38 & 16.28 & 16.50 & 21.89 & 21.23 & 20.10 \\

SEGS-SLAM & 19.43 & 19.81 & 18.44 & 21.75 & 17.33 & - & 18.44 & 15.37 & 20.68 & 19.54 & 26.14 \\

S3PO-GS & 16.70 & 20.09 & 18.52 & 22.69 & 18.41 & 17.14 & 16.67 & 19.02 & 22.78 & 23.06 & 20.74 \\

\midrule

OnTheFly-NVS & 18.44 & 18.91 & 18.38 & 25.43 & 15.84 & 17.26 & 20.86 & 16.27 & 25.69 & 20.04 & 19.80 \\

LongSplat & \textbf{27.07} & \underline{25.35} & \underline{25.48} & \underline{28.14} & \underline{21.27} & \underline{23.52} & \underline{22.81} & \underline{22.36} & \textbf{26.75} & \underline{27.46} & \underline{25.80} \\

\midrule

Ours & \underline{26.19} & \textbf{26.04} & \textbf{25.54} & \textbf{29.43} & \textbf{24.06} & \textbf{25.23} & \textbf{24.92} & \textbf{23.30} & \underline{26.50} & \textbf{29.91} & \textbf{26.92} \\
\bottomrule
\end{tabular}%
}
\end{center}
\end{table}

\begin{table}[htbp]
\begin{center}
\renewcommand{\arraystretch}{1.15}
\setlength{\tabcolsep}{2pt}
\footnotesize
\caption{SSIM on the TUM Dataset}   
\label{tab:ssim_tum}%
\resizebox{1\linewidth}{!}{
\begin{tabular}{c|ccccccccccc}
\toprule
Method & f1\_360 & f1\_desk & f1\_desk2 & f1\_floor & f1\_plant & f1\_room & f1\_rpy & f1\_teddy & f1\_xyz & f2\_xyz & f3\_office \\
\midrule
MonoGS & 0.583 & 0.529 & 0.552 & 0.586 & 0.581 & 0.542 & 0.575 & 0.539 & 0.738 & 0.698 & 0.698 \\

SEGS-SLAM & 0.751 & 0.775 & 0.720 & 0.752 & 0.641 & - & 0.718 & 0.622 & 0.821 & 0.769 & \underline{0.861} \\

S3PO-GS & 0.602 & 0.680 & 0.650 & 0.625 & 0.616 & 0.597 & 0.596 & 0.614 & 0.762 & 0.752 & 0.723 \\

\midrule

OnTheFly-NVS & 0.725 & 0.712 & 0.713 & \underline{0.783} & 0.600 & 0.662 & 0.755 & 0.594 & 0.870 & 0.730 & 0.760 \\

LongSplat & \underline{0.826} & \underline{0.833} & \underline{0.842} & 0.767 & \underline{0.695} & \underline{0.790} & \underline{0.783} & \underline{0.736} & \textbf{0.888} & \underline{0.879} & 0.799 \\

\midrule

Ours & \textbf{0.850} & \textbf{0.861} & \textbf{0.859} & \textbf{0.838} & \textbf{0.797} & \textbf{0.838} & \textbf{0.847} & \textbf{0.768} & \underline{0.883} & \textbf{0.922} & \textbf{0.882} \\
\bottomrule
\end{tabular}%
}
\end{center}
\end{table}

\begin{table}[htbp]
\begin{center}
\renewcommand{\arraystretch}{1.15}
\setlength{\tabcolsep}{2pt}
\footnotesize
\caption{LPIPS on the TUM Dataset}   
\label{tab:lpips_tum}%
\resizebox{1\linewidth}{!}{
\begin{tabular}{c|ccccccccccc}
\toprule
Method & f1\_360 & f1\_desk & f1\_desk2 & f1\_floor & f1\_plant & f1\_room & f1\_rpy & f1\_teddy & f1\_xyz & f2\_xyz & f3\_office \\
\midrule
MonoGS & 0.642 & 0.664 & 0.661 & 0.736 & 0.572 & 0.679 & 0.509 & 0.654 & 0.326 & 0.355 & 0.511 \\

SEGS-SLAM & 0.361 & \underline{0.244} & 0.358 & 0.277 & \underline{0.399} & - & 0.336 & 0.415 & 0.205 & 0.270 & 0.200 \\

S3PO-GS & 0.551 & 0.433 & 0.506 & 0.634 & 0.461 & 0.571 & 0.507 & 0.482 & 0.296 & 0.270 & 0.420 \\

\midrule

OnTheFly-NVS & 0.445 & 0.396 & 0.404 & \underline{0.305} & 0.501 & 0.440 & 0.353 & 0.505 & 0.199 & 0.326 & \underline{0.307} \\

LongSplat & \underline{0.324} & 0.266 & \underline{0.270} & \underline{0.305} & 0.406 & \underline{0.255} & \underline{0.267} & \underline{0.300} & \textbf{0.127} & \underline{0.152} & 0.325 \\

\midrule

Ours & \textbf{0.279} & \textbf{0.220} & \textbf{0.238} & \textbf{0.233} & \textbf{0.263} & \textbf{0.251} & \textbf{0.235} & \textbf{0.298} & \underline{0.174} & \textbf{0.080} & \textbf{0.191} \\
\bottomrule
\end{tabular}%
}
\end{center}
\end{table}
\begin{table}[htbp]
\begin{center}
\renewcommand{\arraystretch}{1.15}
\setlength{\tabcolsep}{4pt}
\footnotesize
\caption{PSNR on the ScanNet Dataset}   
\label{tab:psnr_scannet}%
\resizebox{1\linewidth}{!}{
\begin{tabular}{c|cccccc}
\toprule
Method & scene0000\_00 & scene0059\_00 & scene0106\_00 & scene0169\_00 & scene0181\_00 & scene0207\_00 \\
\midrule
MonoGS & - & 18.09 & 18.03 & 19.71 & 19.37 & 19.16 \\

SEGS-SLAM & - & - & 17.92 & 20.89 & \underline{21.40} & 18.69 \\

S3PO-GS & \underline{17.98} & \underline{19.54} & \underline{20.27} & \underline{21.24} & 21.20 & \underline{20.62} \\

\midrule

OnTheFly-NVS & 14.88 & 16.33 & 14.66 & 16.37 & 15.47 & 14.47 \\

LongSplat & - & 19.52 & 19.01 & 19.71 & 18.94 & 19.17 \\

\midrule

Ours & \textbf{23.28} & \textbf{24.74} & \textbf{26.34} & \textbf{23.07} & \textbf{21.80} & \textbf{25.37} \\
\bottomrule
\end{tabular}%
}
\end{center}
\end{table}

\begin{table}[htbp]
\begin{center}
\renewcommand{\arraystretch}{1.15}
\setlength{\tabcolsep}{4pt}
\footnotesize
\caption{SSIM on the ScanNet Dataset}   
\label{tab:ssim_scannet}%
\resizebox{1\linewidth}{!}{
\begin{tabular}{c|cccccc}
\toprule
Method & scene0000\_00 & scene0059\_00 & scene0106\_00 & scene0169\_00 & scene0181\_00 & scene0207\_00 \\
\midrule
MonoGS & - & 0.733 & 0.774 & 0.792 & 0.823 & 0.776 \\

SEGS-SLAM & - & - & 0.804 & \underline{0.850} & \textbf{0.902} & \underline{0.799} \\

S3PO-GS & 0.738 & \underline{0.769} & \underline{0.816} & 0.815 &0.845 & 0.797 \\

\midrule

OnTheFly-NVS & \underline{0.742} & 0.747 & 0.667 & 0.721 & 0.667 & 0.705 \\

LongSplat & - & 0.752 & 0.764 & 0.750 & 0.782 & 0.721 \\

\midrule

Ours & \textbf{0.824} & \textbf{0.863} & \textbf{0.905} & \textbf{0.859} & \underline{0.889} & \textbf{0.848} \\
\bottomrule
\end{tabular}%
}
\end{center}
\end{table}

\begin{table}[htbp]
\begin{center}
\renewcommand{\arraystretch}{1.15}
\setlength{\tabcolsep}{4pt}
\footnotesize
\caption{LPIPS on the ScanNet Dataset}   
\label{tab:lpips_scannet}%
\resizebox{1\linewidth}{!}{
\begin{tabular}{c|cccccc}
\toprule
Method & scene0000\_00 & scene0059\_00 & scene0106\_00 & scene0169\_00 & scene0181\_00 & scene0207\_00 \\
\midrule
MonoGS & - & 0.713 & 0.597 & 0.612 & 0.578 & 0.644 \\

SEGS-SLAM & - & - & 0.405 & \underline{0.352} & \textbf{0.269} & 0.432 \\

S3PO-GS & 0.700 & 0.518 & 0.484 & 0.550 & 0.513 & 0.583 \\

\midrule

OnTheFly-NVS & \underline{0.464} & 0.477 & 0.554 & 0.504 & 0.491 & 0.474 \\

LongSplat & - & \underline{0.402} & \underline{0.384} & 0.384 &0.423 & \underline{0.427} \\

\midrule

Ours & \textbf{0.254} & \textbf{0.279} & \textbf{0.237} & \textbf{0.278} & \underline{0.288} & \textbf{0.290} \\
\bottomrule
\end{tabular}%
}
\end{center}
\end{table}
\begin{table}[htbp]
\begin{center}
\renewcommand{\arraystretch}{1.15}
\setlength{\tabcolsep}{3pt}
\footnotesize
\caption{PSNR on the Waymo Dataset}   
\label{tab:psnr_scannetpp}%
\resizebox{1\linewidth}{!}{
\begin{tabular}{c|ccccccccc}
\toprule
Method & 100613 & 106762 & 132384 & 13476 & 152706 & 153495 & 158686 & 163453 & 405841 \\
\midrule
MonoGS & 20.05 & 20.91 & 22.71 & 19.51 & 21.23 & 14.15 & 20.29 & 19.01 & 16.19 \\

SEGS-SLAM & 20.14 & 23.60 & 22.52 & - & 24.11 & 21.16 & 21.22 & - & 19.01 \\

S3PO-GS & 25.36 & \underline{28.23} & \underline{27.01} & \underline{24.85} & \underline{28.53} & \underline{26.54} & 26.03 &23.65 & \underline{27.28} \\

\midrule

OnTheFly-NVS & \underline{26.95} & 27.21 & 25.31 & 24.34 & 25.45 & 26.41 & \underline{26.30} & \underline{23.97} & 23.79 \\

LongSplat & 24.25 & 23.70 & 24.11 & 24.42 & 25.66 & 23.74 & 24.84 & 22.69 & 25.61 \\

\midrule

Ours & \textbf{27.98} & \textbf{30.84} & \textbf{30.32} & \textbf{28.03} & \textbf{29.80} & \textbf{27.60} & \textbf{26.83} & \textbf{26.91} & \textbf{30.48} \\
\bottomrule
\end{tabular}%
}
\end{center}
\end{table}

\begin{table}[htbp]
\begin{center}
\renewcommand{\arraystretch}{1.15}
\setlength{\tabcolsep}{3pt}
\footnotesize
\caption{SSIM on the Waymo Dataset}   
\label{tab:ssim_scannetpp}%
\resizebox{1\linewidth}{!}{
\begin{tabular}{c|ccccccccc}
\toprule
Method & 100613 & 106762 & 132384 & 13476 & 152706 & 153495 & 158686 & 163453 & 405841 \\
\midrule
MonoGS & 0.758 & 0.816 & 0.855 & 0.713 & 0.792 & 0.672 & 0.723 & 0.745 & 0.695 \\

SEGS-SLAM & 0.730 & 0.805 & 0.824 & - & 0.784 & 0.734 & 0.677 & - & 0.698 \\

S3PO-GS & 0.828 & \underline{0.878} & \underline{0.883} & \underline{0.778} & \underline{0.856} & 0.846 & 0.819 & \underline{0.797} & \underline{0.865} \\

\midrule

OnTheFly-NVS & \underline{0.850} & 0.854 & 0.864 & 0.764 & 0.795 & \underline{0.847} & \underline{0.837} & 0.785 & 0.788 \\

LongSplat & 0.773 & 0.766 & 0.822 & 0.716 & 0.778 & 0.755 & 0.777 & 0.732 & 0.795 \\

\midrule

Ours & \textbf{0.865} & \textbf{0.906} & \textbf{0.919} & \textbf{0.856} & \textbf{0.882} & \textbf{0.871} & \textbf{0.847} & \textbf{0.866} & \textbf{0.907} \\
\bottomrule
\end{tabular}%
}
\end{center}
\end{table}

\begin{table}[htbp]
\begin{center}
\renewcommand{\arraystretch}{1.15}
\setlength{\tabcolsep}{3pt}
\footnotesize
\caption{LPIPS on the Waymo Dataset}   
\label{tab:lpips_scannetpp}%
\resizebox{1\linewidth}{!}{
\begin{tabular}{c|ccccccccc}
\toprule
Method & 100613 & 106762 & 132384 & 13476 & 152706 & 153495 & 158686 & 163453 & 405841 \\
\midrule
MonoGS & 0.610 & 0.534 & 0.451 & 0.745 & 0.666 & 0.710 & 0.627 & 0.664 & 0.633 \\

SEGS-SLAM & 0.484 & 0.399 & 0.384 & - & 0.450 & 0.472 & 0.495 & - & 0.502 \\

S3PO-GS &0.329 & \underline{0.276} & \underline{0.275} & 0.471 & 0.427 & 0.379 & 0.373 & 0.411 & 0.352 \\

\midrule

OnTheFly-NVS & \underline{0.328} &0.337 &0.361 & 0.378 & 0.401 & \underline{0.348} & \underline{0.307} & 0.376 & 0.400 \\

LongSplat & 0.356 & 0.328 & 0.355 & \underline{0.354} & \underline{0.397} & 0.430 & 0.321 & \underline{0.371} & \underline{0.326} \\

\midrule

Ours & \textbf{0.308} & \textbf{0.237} & \textbf{0.265} & \textbf{0.267} & \textbf{0.304} & \textbf{0.313} & \textbf{0.283} & \textbf{0.289} & \textbf{0.216} \\
\bottomrule
\end{tabular}%
}
\end{center}
\end{table}

\begin{table}[htbp]
\begin{center}
\renewcommand{\arraystretch}{1.15}
\setlength{\tabcolsep}{1pt}
\footnotesize
\caption{PSNR on the VR-NeRF Dataset}   
\label{tab:psnr_indoor}
\resizebox{1\linewidth}{!}{
\begin{tabular}{c|cccccccc}
\toprule
Method & appartment262 & kitchen261 & kitchen262 & kitchen263 & table61 & workspace61 & workspace62 & workspace64 \\
\midrule
MonoGS & 18.43 & 16.91 & 11.66 & 14.47 & 15.50 & 14.80 & 15.12 & 14.80 \\
SEGS-SLAM & 26.14 & \textbf{31.81} & - & \textbf{32.65} & \textbf{36.55} & \textbf{30.95} & - & - \\
S3PO-GS & 28.45 & 27.98 & 25.16 & 18.56 & 19.36 & 22.16 & 23.49 & 22.17 \\
\midrule
LongSplat & \underline{31.22} & 27.10 & 27.76 & 24.04 & 24.90 & - & 22.32 & 22.84 \\
OnTheFly-NVS & 30.52 & 30.24 & 27.51 & 25.16 & 27.77 & 22.08 & \textbf{26.04} & \textbf{29.05} \\
\midrule
Ours & \textbf{32.98} & \underline{30.90} & \textbf{30.23} & \underline{29.04} & \underline{29.05} & \underline{24.68} & \underline{24.63} & \underline{27.13} \\
\bottomrule
\end{tabular}}
\end{center}
\end{table}

\begin{table}[htbp]
\begin{center}
\renewcommand{\arraystretch}{1.15}
\setlength{\tabcolsep}{1pt}
\footnotesize
\caption{SSIM on the VR-NeRF Dataset}   
\label{tab:ssim_indoor}
\resizebox{1\linewidth}{!}{
\begin{tabular}{c|cccccccc}
\toprule
Method & appartment262 & kitchen261 & kitchen262 & kitchen263 & table61 & workspace61 & workspace62 & workspace64 \\
\midrule
MonoGS & 0.646 & 0.627 & 0.506 & 0.599 & 0.595 & 0.526 & 0.581 & 0.522 \\
SEGS-SLAM & 0.831 & \underline{0.910} & - & \underline{0.883} & \textbf{0.949} & \textbf{0.905} & - & - \\
S3PO-GS & 0.875 & 0.880 & 0.856 & 0.719 & 0.737 & 0.762 & 0.791 & 0.758 \\
\midrule
LongSplat & 0.905 & 0.861 & 0.888 & 0.803 & 0.823 & - & 0.757 & 0.787 \\
OnTheFly-NVS & \underline{0.912} & 0.903 & \underline{0.900} & 0.847 & 0.882 & 0.775 & \textbf{0.855} & \textbf{0.898} \\
\midrule
Ours & \textbf{0.937} & \textbf{0.913} & \textbf{0.939} & \textbf{0.913} & \underline{0.900} & \underline{0.842} & \underline{0.833} & \underline{0.883} \\
\bottomrule
\end{tabular}}
\end{center}
\end{table}

\begin{table}[htbp]
\begin{center}
\renewcommand{\arraystretch}{1.15}
\setlength{\tabcolsep}{1pt}
\footnotesize
\caption{LPIPS on the VR-NeRF Dataset}   
\label{tab:lpips_indoor}
\resizebox{1\linewidth}{!}{
\begin{tabular}{c|cccccccc}
\toprule
Method & appartment262 & kitchen261 & kitchen262 & kitchen263 & table61 & workspace61 & workspace62 & workspace64 \\
\midrule
MonoGS & 0.631 & 0.688 & 0.676 & 0.662 & 0.582 & 0.689 & 0.721 & 0.685 \\
SEGS-SLAM & 0.375 & \textbf{0.218} & - & \textbf{0.192} & \textbf{0.174} & \textbf{0.203} & - & - \\
S3PO-GS & 0.321 & 0.302 & 0.269 & 0.597 & 0.474 & 0.385 & 0.350 & 0.386 \\
\midrule
LongSplat & \underline{0.292} & 0.308 & \underline{0.265} & 0.319 & 0.353 & - & 0.359 & 0.352 \\
OnTheFly-NVS & 0.302 & 0.277 & 0.311 & 0.311 & 0.322 & 0.385 & \textbf{0.307} & \textbf{0.261} \\
\midrule
Ours & \textbf{0.201} & \underline{0.224} & \textbf{0.185} & \underline{0.198} & \underline{0.256} & \underline{0.286} & \underline{0.314} & \underline{0.274} \\
\bottomrule
\end{tabular}}
\end{center}
\end{table}

\begin{table}[htbp]
\begin{center}
\renewcommand{\arraystretch}{1.15}
\setlength{\tabcolsep}{9pt}
\footnotesize
\caption{PSNR on the KITTI Dataset}   
\label{tab:psnr_cofusion_KITTI}%
\resizebox{1\linewidth}{!}{
\begin{tabular}{c|cccccccc}
\toprule
Method & 00 & 02 & 03 & 05 & 06 & 07 & 08 & 10 \\
\midrule
MonoGS & 16.01 & 15.08 & 16.90 & 15.66 & 16.38 & 11.38 & 13.21 & 11.82 \\

SEGS-SLAM & 12.69 & 14.88 & 16.71 & 14.88 & 14.65 & 10.45 & 14.50 & 13.51 \\

S3PO-GS & \underline{20.77} & \underline{19.30} & \underline{20.51} & \underline{20.73} & \underline{20.42} & \underline{20.38} & \underline{20.27} & \underline{17.41} \\

\midrule

OnTheFly-NVS & 15.97 & 17.00 & 17.34 & 18.08 & 17.29 & 18.21 & 16.50 & 14.73 \\

LongSplat & 17.84 & 14.62 & 17.97 & 18.08 & 18.68 & 16.14 & 16.58 & 14.93 \\

\midrule

Ours & \textbf{23.76} & \textbf{22.53} & \textbf{24.54} & \textbf{23.80} & \textbf{23.59} & \textbf{23.92} & \textbf{22.86} & \textbf{20.38} \\
\bottomrule
\end{tabular}%
}
\end{center}
\end{table}

\begin{table}[htbp]
\begin{center}
\renewcommand{\arraystretch}{1.15}
\setlength{\tabcolsep}{9pt}
\footnotesize
\caption{SSIM on the KITTI Dataset}   
\label{tab:ssim_cofusion_KITTI}%
\resizebox{1\linewidth}{!}{
\begin{tabular}{c|cccccccc}
\toprule
Method & 00 & 02 & 03 & 05 & 06 & 07 & 08 & 10 \\
\midrule
MonoGS & 0.568 & 0.476 & 0.487 & 0.491 & 0.560 & 0.439 & 0.480 & 0.420 \\

SEGS-SLAM & 0.454 & 0.458 & 0.448 & 0.470 & 0.510 & 0.405 & 0.493 & 0.462 \\

S3PO-GS & \underline{0.732} & \underline{0.587} & \underline{0.581} & \underline{0.659} & \underline{0.652} & \underline{0.715} & \underline{0.684} & \underline{0.545} \\

\midrule

OnTheFly-NVS & 0.594 & 0.538 & 0.529 & 0.606 & 0.583 & 0.673 & 0.622 & 0.483 \\

LongSplat & 0.616 & 0.444 & 0.482 & 0.548 & 0.583 & 0.557 & 0.543 & 0.484 \\

\midrule

Ours & \textbf{0.829} & \textbf{0.707} & \textbf{0.745} & \textbf{0.781} & \textbf{0.779} & \textbf{0.827} & \textbf{0.786} & \textbf{0.663} \\
\bottomrule
\end{tabular}%
}
\end{center}
\end{table}

\begin{table}[htbp]
\begin{center}
\renewcommand{\arraystretch}{1.15}
\setlength{\tabcolsep}{9pt}
\footnotesize
\caption{LPIPS on the KITTI Dataset}   
\label{tab:lpips_cofusion_KITTI}%
\resizebox{1\linewidth}{!}{
\begin{tabular}{c|cccccccc}
\toprule
Method & 00 & 02 & 03 & 05 & 06 & 07 & 08 & 10 \\
\midrule
MonoGS & 0.687 & 0.761 & 0.735 & 0.753 & 0.720 & 0.826 & 0.830 & 0.820 \\

SEGS-SLAM & 0.492 & 0.491 & 0.465 & 0.453 & 0.464 & 0.552 & 0.450 & 0.533 \\

S3PO-GS & \underline{0.264} & \underline{0.459} & 0.498 & \underline{0.389} & 0.401 & \underline{0.364} & \underline{0.345} & 0.563 \\

\midrule

OnTheFly-NVS & 0.461 & 0.501 & 0.501 & 0.455 & 0.500 & 0.388 & 0.422 & 0.540 \\

LongSplat & 0.375 & 0.531 & \underline{0.438} & 0.428 & \underline{0.394} & 0.454 & 0.440 & \underline{0.517} \\

\midrule

Ours & \textbf{0.234} & \textbf{0.375} & \textbf{0.311} & \textbf{0.281} & \textbf{0.288} & \textbf{0.242} & \textbf{0.266} & \textbf{0.394} \\
\bottomrule
\end{tabular}%
}
\end{center}
\end{table}

\begin{table}[htbp]
\begin{center}
\renewcommand{\arraystretch}{1.15}
\setlength{\tabcolsep}{6pt}
\footnotesize
\caption{PSNR on the ScanNet++ Dataset}   
\label{tab:track_scannetpp_psnr}%
\resizebox{1\linewidth}{!}{
\begin{tabular}{c|ccccccc}
\toprule
Method & 00777c41d4 & 02f25e5fee & 0b031f3119 & 126d03d821 & 1cbb105c6a & 2284bf5c9d & 2d2e873aa0 \\
\midrule
MonoGS & 14.621 & 20.915 & 9.488 & 21.353 & 23.716 & 13.049 & 16.861 \\
SEGS-SLAM & - & \underline{28.177} & - & - & - & - & - \\
S3PO-GS & \underline{21.316} & 25.158 & 21.019 & 23.758 & 26.614 & \underline{23.446} & \underline{23.554} \\
OnTheFly-NVS & 16.270 & 19.134 & 21.701 & 17.572 & 17.478 & 18.200 & 14.173 \\
LongSplat & 18.465 & 26.747 & \underline{22.484} & \underline{27.251} & \underline{28.621} & 23.348 & - \\
Ours & \textbf{26.543} & \textbf{30.671} & \textbf{27.722} & \textbf{32.796} & \textbf{32.406} & \textbf{30.277} & \textbf{28.137} \\
\midrule
Method & 303745abc7 & 41eb967018 & 46001f434d & 4808c4a397 & 546292a9db & 712dc47104 & 7543973e1a \\
\midrule
MonoGS & 14.899 & 18.011 & 18.515 & 21.570 & 9.464 & 8.970 & 22.547 \\
SEGS-SLAM & - & - & - & - & - & - & 27.169 \\
S3PO-GS & \underline{25.840} & \underline{21.074} & 20.288 & \underline{24.263} & 20.612 & \underline{17.843} & 26.367 \\
OnTheFly-NVS & 15.884 & 16.247 & 16.426 & 18.722 & \underline{24.791} & 14.056 & 21.540 \\
LongSplat & - & - & \textbf{24.437} & - & 24.587 & - & \underline{28.539} \\
Ours & \textbf{33.056} & \textbf{28.608} & \underline{21.716} & \textbf{30.101} & \textbf{25.868} & \textbf{27.506} & \textbf{32.308} \\
\bottomrule
\end{tabular}%
}
\end{center}
\end{table}

\begin{table}[htbp]
\begin{center}
\renewcommand{\arraystretch}{1.15}
\setlength{\tabcolsep}{6pt}
\footnotesize
\caption{SSIM on the ScaNnet++ Dataset}   
\label{tab:track_scannetpp_ssim}%
\resizebox{1\linewidth}{!}{
\begin{tabular}{c|ccccccc}
\toprule
Method & 00777c41d4 & 02f25e5fee & 0b031f3119 & 126d03d821 & 1cbb105c6a & 2284bf5c9d & 2d2e873aa0 \\
\midrule
MonoGS & 0.559 & 0.784 & 0.485 & 0.796 & 0.833 & 0.622 & 0.641 \\
SEGS-SLAM & - & \underline{0.911} & - & - & - & - & - \\
S3PO-GS & \underline{0.695} & 0.853 & 0.809 & 0.828 & 0.876 & \underline{0.833} & \underline{0.830} \\
OnTheFly-NVS & 0.583 & 0.779 & \underline{0.823} & 0.753 & 0.749 & 0.751 & 0.705 \\
LongSplat & 0.604 & 0.868 & 0.816 & \underline{0.854} & \underline{0.897} & 0.819 & - \\
Ours & \textbf{0.855} & \textbf{0.941} & \textbf{0.903} & \textbf{0.942} & \textbf{0.956} & \textbf{0.937} & \textbf{0.918} \\
\midrule
Method & 303745abc7 & 41eb967018 & 46001f434d & 4808c4a397 & 546292a9db & 712dc47104 & 7543973e1a \\
\midrule
MonoGS & 0.723 & 0.737 & 0.844 & 0.848 & 0.244 & 0.590 & 0.836 \\
SEGS-SLAM & - & - & - & - & - & - & 0.891 \\
S3PO-GS & \underline{0.878} & \underline{0.788} & 0.861 & \underline{0.873} & 0.688 & \underline{0.784} & 0.884 \\
OnTheFly-NVS & 0.780 & 0.705 & 0.817 & 0.826 & \underline{0.825} & 0.732 & 0.831 \\
LongSplat & - & - & \textbf{0.900} & - & 0.793 & - & \underline{0.894} \\
Ours & \textbf{0.959} & \textbf{0.905} & \underline{0.891} & \textbf{0.938} & \textbf{0.850} & \textbf{0.915} & \textbf{0.950} \\
\bottomrule
\end{tabular}%
}
\end{center}
\end{table}

\begin{table}[htbp]
\begin{center}
\renewcommand{\arraystretch}{1.15}
\setlength{\tabcolsep}{6pt}
\footnotesize
\caption{LPIPS on the ScanNet++ Dataset}   
\label{tab:track_scannetpp_lpips}%
\resizebox{1\linewidth}{!}{
\begin{tabular}{c|ccccccc}
\toprule
Method & 00777c41d4 & 02f25e5fee & 0b031f3119 & 126d03d821 & 1cbb105c6a & 2284bf5c9d & 2d2e873aa0 \\
\midrule
MonoGS & 0.822 & 0.459 & 0.724 & 0.450 & 0.370 & 0.710 & 0.769 \\
SEGS-SLAM & - & \underline{0.148} & - & - & - & - & - \\
S3PO-GS & 0.451 & 0.243 & 0.350 & 0.346 & 0.264 & 0.289 & \underline{0.316} \\
OnTheFly-NVS & 0.524 & 0.352 & 0.310 & 0.404 & 0.409 & 0.386 & 0.499 \\
LongSplat & \underline{0.436} & 0.179 & \underline{0.307} & \underline{0.242} & \underline{0.179} & \underline{0.245} & - \\
Ours & \textbf{0.209} & \textbf{0.122} & \textbf{0.196} & \textbf{0.143} & \textbf{0.123} & \textbf{0.134} & \textbf{0.179} \\
\midrule
Method & 303745abc7 & 41eb967018 & 46001f434d & 4808c4a397 & 546292a9db & 712dc47104 & 7543973e1a \\
\midrule
MonoGS & 0.707 & 0.619 & 0.485 & 0.408 & 0.734 & 0.707 & 0.441 \\
SEGS-SLAM & - & - & - & - & - & - & 0.216 \\
S3PO-GS & \underline{0.259} & 0.454 & 0.419 & \underline{0.300} & 0.520 & \underline{0.449} & 0.303 \\
OnTheFly-NVS & 0.382 & \underline{0.434} & 0.362 & 0.331 & \underline{0.256} & 0.450 & 0.305 \\
LongSplat & - & - & \textbf{0.265} & - & 0.286 & - & \underline{0.200} \\
Ours & \textbf{0.115} & \textbf{0.165} & \underline{0.279} & \textbf{0.156} & \textbf{0.205} & \textbf{0.176} & \textbf{0.140} \\
\bottomrule
\end{tabular}%
}
\end{center}
\end{table}


\begin{table}[htbp]
\begin{center}
\renewcommand{\arraystretch}{1.15}
\setlength{\tabcolsep}{2pt}
\footnotesize
\caption{Tracking Results on the ScanNet++ Dataset}
\label{tab:track_scannetpp}%
\resizebox{1\linewidth}{!}{
\begin{tabular}{c|ccccccccccccccc}
\toprule
   Method   &0077  &02f2  &0b03  &126d  &1cbb  &2284  &2d2e  &3037  &41eb  &4600  &4808  &5462  &712d  &7543 \\
\midrule
OnTheFly    &0.804 &0.157 &0.114 &0.227 &0.227 &0.913 &1.438 &1.186 &1.605 &3.932 &0.486 &\textbf{0.018} &0.458 &0.907  \\
LongSplat   &1.307 &0.041 &0.378 &0.550 &0.023 &0.713 &-     &-     &-     &1.943 &-     &0.408 &-     &0.058 \\
\midrule
MonoGS      &1.364 &0.626 &-     &-     &0.294 &0.939 &2.019 &0.922 &1.760 &3.540 &0.432 &-     &0.853 &0.642    \\
S3PO-GS     &0.650 &0.016 &0.093 &0.418 &0.135 &0.441 &0.355 &0.858 &1.674 &1.318 &0.614 &0.909 &1.092 &0.270    \\
SEGS-SLAM   &-     &\textbf{0.005} &-     &-     &-     &-     &-     &-     &-     &-     &-     &-     &-     &0.485    \\
MASt3R-SLAM &0.121 &0.017 &0.059 &\underline{0.122} &0.021 &0.027 &0.835 &0.176 &\underline{0.020} &2.379 &0.065 &0.042 &0.021 &0.025   \\
loop with vggt   &\underline{0.010} &\underline{0.011} &\textbf{0.014} &1.176 &\textbf{0.014} &\underline{0.022} &\textbf{0.010} &\underline{0.007} &0.030 &\underline{0.075} &\underline{0.022} &0.032 &\textbf{0.013} &\textbf{0.011} \\ 
Ours  &\textbf{0.009} &\underline{0.011} &\underline{0.015} &\textbf{0.019} &\underline{0.015} &\textbf{0.012} &\textbf{0.010} &\textbf{0.005} &\textbf{0.017} &\textbf{0.060} &\textbf{0.021} &\underline{0.030} &\underline{0.014} &\textbf{0.011}   \\

\bottomrule
\end{tabular}%
}
\end{center}
\end{table}%

\begin{table}[htbp]
\begin{center}
\renewcommand{\arraystretch}{1.15}
\setlength{\tabcolsep}{2pt}
\footnotesize
\caption{Tracking Results on the TUM Dataset}
\label{tab:track_tum}%
\resizebox{1\linewidth}{!}{
\begin{tabular}{c|ccccccccccc}
\toprule
Method & f1\_360 & f1\_desk & f1\_desk2 & f1\_floor & f1\_plant & f1\_room & f1\_rpy & f1\_teddy & f1\_xyz & f2\_xyz & f3\_office \\
\midrule
OnTheFly    &0.187 &-     &-     &-     &-     &0.874 &4.136 &-     &0.278 &0.073 &1.489        \\
LongSplat   &0.133 &-     &-     &-     &-     &0.712 &-     &-     &-     &0.103 &-            \\                              
\midrule
MonoGS      &0.160 &0.034 &0.596 &0.531 &0.077 &0.649 &0.032 &0.512 &0.017 &0.047 &0.033   \\
S3PO-GS     &0.093 &0.041 &0.155 &0.228 &0.040 &0.552 &0.053 &0.051 &0.010 &0.016 &0.045   \\
SEGS-SLAM   &0.154 &\textbf{0.016} &\textbf{0.013} &-     &-     &-     &-     &0.293 &0.009 &0.006 &0.026   \\
MASt3R-SLAM &0.049 &\textbf{0.016} &\underline{0.024} &\textbf{0.025} &0.020 &0.061 &0.027 &\textbf{0.041} &0.009 &0.005 &0.031   \\
loop with vggt     &\textbf{0.040} &\textbf{0.016} &0.026 &0.026 &\underline{0.017} &\textbf{0.056} &\underline{0.023} &0.047 &\textbf{0.007} &\textbf{0.005} &\underline{0.024} \\    
Ours      &\textbf{0.040} &\textbf{0.016} &0.025 &\textbf{0.025} &\textbf{0.016} &\underline{0.060} &\textbf{0.022} &\underline{0.045} &\textbf{0.007} &\textbf{0.005} &\textbf{0.019}   \\  %

\bottomrule
\end{tabular}%
}
\end{center}
\end{table}%

\begin{table}[htbp]
\begin{center}
\footnotesize
\caption{Tracking Results on the KITTI Dataset}
\label{tab:track_kitti}%
\resizebox{1\linewidth}{!}{
\begin{tabular}{c|ccccccccc}
\toprule
   Method   & 00 &02 &03 &05 &06 &07 &08 &10 &Avg. \\
\midrule
OnTheFly    &18.297 &28.230 &14.356 &5.027 &9.608 &1.550 &9.836 &9.502 &12.051 \\
LongSplat   &3.047  &10.796 &9.344  &9.482 &6.864 &2.866 &4.480 &6.576 &6.682  \\
\midrule
MonoGS      &11.868 &11.817 &11.195 &4.623 &7.638 &4.128 &5.864 &5.109 &7.780 \\
S3PO-GS     &\underline{1.196}  &2.961  &5.522  &1.459 &\textbf{0.721} &\underline{1.009} &2.655 &1.927 &2.181 \\
SEGS-SLAM   &\textbf{0.561}  &\textbf{0.897}  &\textbf{0.189}  &\underline{0.967} &4.122 &\textbf{0.851} &\textbf{0.477} &\textbf{0.353} &\textbf{1.052} \\
MASt3R-SLAM &-      &\underline{1.756}  &\underline{0.397}  &\textbf{0.761} &2.279 &1.361 &-     &-     & --   \\
loop with vggt        &1.304 & 2.691 & 0.442 & 1.103 & \underline{0.958} & 1.257 & \underline{1.160} & \underline{1.893} & \underline{1.351} \\
Ours        &1.304  &2.691  &0.442  &1.103 &\underline{0.958} &1.321 &1.167 &\underline{1.893} &1.360 \\

\bottomrule
\end{tabular}%
}
\end{center}
\end{table}%

\begin{table}[htbp]
\begin{center}
    \footnotesize
    \caption{Tracking Results on the Waymo Dataset}
    \label{tab:track_waymo}%
    \resizebox{1\linewidth}{!}{
    \begin{tabular}{c|cccccccccc}
    \toprule
       Method   & 13476 &100613 &106762 &132384 &152706 &153495 &158686 &163453 &405841 &Avg. \\
    \midrule
    OnTheFly    &1.687 &1.051 &2.022 &14.395 &1.193 &2.110 &2.206 &1.872 &1.523 &3.118  \\
    LongSplat   &7.024 &10.242&7.506 &2.616  &4.900 &2.625 &3.921 &3.939 &2.729 &4.956  \\
    \midrule
    MonoGS      &3.262 &6.706 &15.638&11.093&8.286 &\textbf{0.585} &6.881 &10.669&3.218 &7.370       \\
    S3PO-GS     &\textbf{1.165} &2.471 &\textbf{0.214} &\textbf{1.551} &1.144 &1.972 &0.963 &1.085 &\textbf{0.560} &1.236      \\
    SEGS-SLAM   &-     &0.860 &\underline{0.696} &2.755  &1.867 &-     &-     &-     &-    &-      \\
    MASt3R-SLAM &-     &-     &2.958 &-     &-     &\underline{1.569} &1.098 &2.334 &0.873 &-      \\
    loop with vggt        &1.230 &\textbf{0.315} &2.762 &\underline{1.753} &\textbf{0.931} &1.578 &\textbf{0.460} &\textbf{1.071} &\underline{0.816} &\textbf{1.213}   \\
    Ours         &\underline{1.229} &\textbf{0.315} &2.762 &\underline{1.753} &\textbf{0.931} &1.578 &\textbf{0.460} &\textbf{1.071} &\underline{0.816} &\textbf{1.213}  \\

    \bottomrule
    \end{tabular}
    }
\end{center}
\end{table}%

\begin{table}[htbp]
\begin{center}
    \footnotesize
    \caption{Tracking Results on the TUM Dataset Compared with Non 3DGS SLAM Systems}
    \label{tab:track_tum_classic}%
    \resizebox{1\linewidth}{!}{
    \begin{tabular}{c|cccccccccc}
    \toprule
       Method   & 360  &desk  &desk2 &floor &plant &room  &rpy   &teddy &xyz   &Avg. \\
    \midrule
    ORB-SLAM3   &-     &\underline{0.017} &0.210 &-     &0.034 &-     &-     &-     &0.009  &-    \\
    DPV-SLAM++  &0.132 &0.018 &0.029 &0.050 &\underline{0.022} &0.096 &0.032 &0.098 &0.010 &0.054\\
    DROID-SLAM  &0.111 &0.018 &0.042 &\textbf{0.021} &\textbf{0.016} &\textbf{0.049} &0.026 &0.048 &0.012 &0.038\\
    Go-SLAM     &0.089 &\textbf{0.016} &0.028 &\underline{0.025} &0.026 &\underline{0.052} &\textbf{0.019} &0.048 &0.010 &0.035\\
    MASt3R-SLAM &\underline{0.049} &\textbf{0.016} &\textbf{0.024} &\underline{0.025} &0.020 &0.061 &0.027 &\textbf{0.041} &\underline{0.009} &\underline{0.030}\\
    Ours        &\textbf{0.040} &\textbf{0.016} &\underline{0.025} &\underline{0.025} &\textbf{0.016} &0.060 &\underline{0.022} &\underline{0.045} &\textbf{0.007} &\textbf{0.028}\\
    \bottomrule
    \end{tabular}%
    }
\end{center}
\end{table}%

\begin{figure*}[t!]
\includegraphics[width=\linewidth]{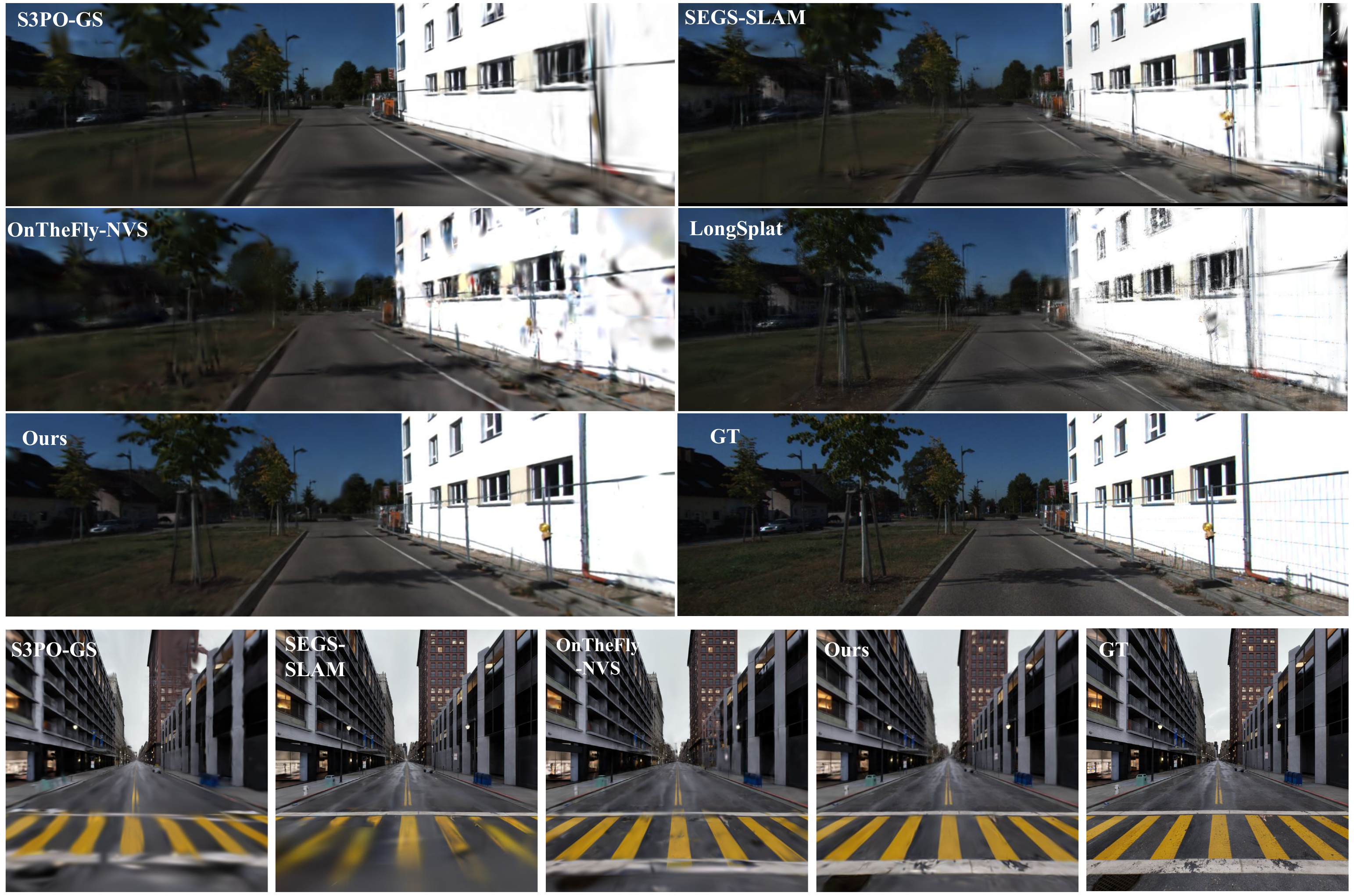}
\caption{More Qualitative Reconstruction Results.}
\label{fig:qualtive_rendering_2}
\end{figure*}

